\RequirePackage{fix-cm}
\documentclass{svjour3}                     
\smartqed  
\usepackage{graphicx}
\usepackage[table,xcdraw]{xcolor}
\usepackage{amsfonts,amsmath}
\usepackage{lineno,hyperref}
\usepackage{subfig}
\usepackage{float}
\usepackage{multirow}
\usepackage[section]{placeins} 
\usepackage[numbers,sort&compress]{natbib}
\usepackage{xcolor} 
\usepackage[section]{placeins} 

\newtheorem{Ex}{Example}
\newcommand{\bp}{{\boldsymbol{p}}}
\newcommand{\bx}{{\boldsymbol{x}}}

\newcommand{\bU}{{\boldsymbol{U}}}
\newcommand{\btU}{{\widetilde{\boldsymbol{U}}}}

\newcommand{\deriv}[2]{\frac{\partial #1}{\partial #2}}

\journalname{Applied Intelligence}
\begin{document}

\title{Evidential fully convolutional network for semantic segmentation
}
\author{Zheng Tong  \and Philippe Xu \and Thierry Den{\oe}ux}
\institute{Zheng Tong, Philippe Xu \at
              Universit\'e de technologie de Compi\`egne, CNRS, Heudiasyc, Compi\`egne, France \\
              \email{zheng.tong@hds.utc.fr, philippe.xu@hds.utc.fr}           
           \and
           Thierry Den{\oe}ux \at
           Universit\'e de technologie de Compi\`egne, CNRS, Heudiasyc, Compi\`egne, France, and\\
             Institut universitaire de France, Paris, France\\
             \email{thierry.denoeux@hds.utc.fr}
}
\date{Received: date / Accepted: date}
\maketitle
\begin{abstract}
We  propose a hybrid architecture composed of  a fully convolutional network (FCN)  and a Dempster-Shafer layer for image semantic segmentation. In the so-called evidential FCN (E-FCN), an encoder-decoder architecture first extracts pixel-wise feature maps from an input image. A Dempster-Shafer layer then computes mass functions at each pixel location based on distances to prototypes. Finally, a utility layer performs semantic segmentation from mass functions and allows for imprecise classification of ambiguous pixels and outliers. We propose an end-to-end learning strategy for jointly updating the network parameters, which can make use of soft (imprecise) labels. Experiments using three databases (Pascal VOC 2011, MIT-scene Parsing and SIFT Flow) show that the proposed combination improves the accuracy and calibration of semantic segmentation by assigning confusing pixels to multi-class sets. 
\end{abstract}
\section{Introduction}
\label{sec:introduction}

In the past few decades, one of the most difficult problems in computer vision has been image semantic segmentation, which is defined as  the process of partitioning a digital image into multiple sets of pixels. The result of image segmentation is a set of segments that collectively cover the entire image, called the \emph{segmentation mask}. The mask constitutes a simplified representation, more meaningful and easier to analyze than the original image. Semantic segmentation has been widely applied to advanced driver assistance systems \cite{xu2016multimodal,ess2009segmentation,cordts2016cityscapes}, human-machine interaction \cite{yoon2015learning}, medical imaging \cite{forouzanfar2010parameter}, and so on.

In the last decade, deep learning-based models, especially fully convolutional networks (FCNs) \cite{long2015fully} and variants \cite{noh2015learning,krahenbuhl2011efficient}, have been developed for semantic segmentation and have achieved remarkable success.  FCNs
 take advantage of existing deep neural networks, which have the capacity to learn reliable and robust features. An FCN transforms  existing and well-known classification models such as  VGG (16-layer net) \cite{simonyan2014very} or ResNet \cite{he2016deep} into fully convolutional ones by replacing
the fully connected layers with convolutional ones to output spatial maps instead of classification scores. Those maps are upsampled using fractionally-strided convolutions (called \emph{deconvolutions}  \cite{zeiler2011adaptive,zeiler2014visualizing}) to produce dense per-pixel labeled outputs. This approach has allowed for significant improvements in segmentation accuracy over traditional methods on benchmark databases like Pascal VOC 2011 \cite{everingham2015pascal}. However, despite the power and flexibility of the FCN-based models, they still face the following three problems:

\begin{enumerate}
 \item \emph{How to perform novelty detection?} In many learning sets, not all classes are labeled, especially for some objects in the background. An ideal image segmentation algorithm should detect ``unknown'' objects belonging to classes that are not represented in the learning set. This capacity is called \emph{novelty detection} \cite{denoeux96b}.  FCN-based models generally randomly assign unknown objects to one of the known classes, though some models tend to assign unknown objects to the background class.
 \item \emph{How to process pixels with confusing information?}  In  image-segmentation training sets, all pixels are precisely labeled, even if the true label is actually uncertain. This is the case, for example, for the pixels at object borders. Pixels with precise but incorrect labels may have negative effects on learning systems \cite{biggio2011support,natarajan2013learning}.
 \item \emph{When will the FCN-based methods fail?}  In decision-making systems, a neural network should not only be as accurate as possible, but it  should also have the ability to indicate when it is likely to be incorrect. Neural networks developed nowadays tend  not to be well calibrated  \cite{guo2017calibration}, though they are more accurate than they were a decade ago. In other words, the accuracy of modern neural networks, including FCN-based models, does not match their confidence.
\end{enumerate}

Dempster-Shafer (DS) theory may provide a  solution to these problems. The DS theory of belief functions \cite{dempster2008upper,shafer1976mathematical}, also referred to as \emph{Evidence Theory}, is based on representing independent pieces of evidence by mass functions and combining them using a generic operator called Dempster's rule. It is a well-established formalism for reasoning and making decisions with uncertainty \cite{denzux201640,denoeux20b,yager2008classic}. A mass function has more degrees of freedom than a probability distribution, which allows it to represent a wider range of belief states, from complete ignorance to full certainty.

One of the applications of the DS theory is to design \emph{evidential classifiers} (e.g., \cite{denoeux2000neural,denoeux19f,lian15,su18}),  which compute a predicted mass function for each input vector. The output mass function can then be used for decision-making \cite{chen2018evidential,denoeux96b,7532858}. Over the years, two main principles have been developed to design an evidential classifier: the model-based and distance-based approaches \cite{denuxdenoeux2006classification}. The former uses estimated class-conditional distributions \cite{smets1993belief}, while the latter constructs mass functions based on distances to prototypes \cite{denoeux2000neural,denoeux19f,lian15}. Thanks to the generality and expressiveness of the belief-function formalism, an evidential classifier provides more informative outputs than those of conventional classifiers (e.g., a neural network with a softmax output layer). The flexibility of evidential classifiers can be exploited for uncertain data classification \cite{yuan2020evidential} and set-valued classification \cite{denoeux96b,MA2021106742}.  Therefore, it may be advantageous to combine an FCN-based model with an evidential classifier for semantic segmentation.

The objective of this study is to take advantage of object representations generated by an FCN and use them as the input features of an evidential classifier for decision-making. The proposed model, referred to as the  \emph{evidential fully convolutional network} (E-FCN), transforms an FCN model by replacing its softmax layer by a distance-based DS layer and a utility layer. In an E-FCN, an FCN model is used to extract pixel-wise high-order features from an input image. Then, the features are converted into pixel-wise mass functions by the DS layer. Finally, the mass functions are used to compute the utilities of acts assigning pixels to a set of classes for semantic segmentation in the proposed utility layer. An end-to-end learning procedure allows us to train the E-FCN using a learning set with soft labels. The effectiveness of the E-FCN  is demonstrated and discussed in the experiments using three benchmark databases (Pascal VOC 2011 \cite{everingham2015pascal}, MIT-scene Parsing \cite{zhou2016semantic}, and SIFT Flow \cite{tighe2010superparsing}).

The rest of the paper is organized as follows. Section \ref{sec:background} starts with a brief reminder of DS theory, the DS layer for constructing mass functions, and feature representation via FCN. The E-FCN model is then introduced in Section \ref{sec:efcn}. Section \ref{sec:experiments} presents numerical experiments, which demonstrate the advantages of the E-FCNs. Finally, we conclude the paper in Section \ref{sec:conclusions}.

\section{Background}
\label{sec:background}

This section first recalls some necessary definitions regarding DS theory (Section \ref{sec:DS_theory}) and the evidential neural network (Section \ref{sec:DS_layer}). A brief description of feature representation via FCNs is then provided in Section \ref{sec:fcn}.

\subsection{Dempster-Shafer theory}
\label{sec:DS_theory}

The main concepts underlying DS theory are only briefly presented in this section, and some basic notations are introduced. Detailed information can be found in Shafer's original work \cite{shafer1976mathematical} and in the recent review \cite{denoeux20b}.

Let $\Omega=\{\omega_1,\ldots,\omega_M\}$ be a set of classes, called the \emph{frame of discernment}. A \emph{mass function} on $\Omega$ is a mapping $m$ from $2^\Omega$ to [0,1] such that $m(\emptyset)=0$ and
\begin{equation}
 \sum_{A\subseteq\Omega}m(A)=1.
\end{equation}
For any $A\subseteq\Omega$, each mass $m(A)$ is interpreted as a share of a unit mass of belief allocated to the hypothesis that the truth is in $A$, and which cannot be allocated to any strict subset of $A$ based on the available evidence. Set $A$ is called a \emph{focal set} of $m$ if $m(A)>0$. A mass function is said to be \emph{logical} if it has only one focal set.  

Two  mass functions $m_1$ and $m_2$ representing independent items of evidence can be combined conjunctively by Dempster's rule $\oplus$ \cite{shafer1976mathematical} as
\begin{subequations}
 \label{con:dempster}
 \begin{equation}
  \label{con:dempster1}
  (m_1\oplus m_2)\left(A\right)=\frac{(m_1 \cap m_2)(A)}{1-(m_1 \cap m_2)(\emptyset)}
 \end{equation}
for all $A\neq\emptyset$, with
 \begin{equation}
  \label{con:dempster2}
  (m_1\cap m_2)(A)=\sum_{B\cap C=A}m_1\left(B\right)m_2\left(C\right),
 \end{equation}
\end{subequations}
and $(m_1\oplus m_2)(\emptyset)=0$. Mass functions $m_1$ and $m_2$ can be combined if and only if the denominator on the right-hand side of Eq. \eqref{con:dempster1} is strictly positive. The operator $\oplus$ is commutative and associative.

For decision-making with belief functions, let $u_{ij} \in [0,1]$ denote the utility of selecting $\omega_i$ when the true state is $\omega_j$,  and $f_{\omega_i}$  the act of selecting $\omega_i$. We define the \emph{pignistic expected utility}  \cite{denoeux2019decision} of act $f_{\omega_i}$ as
\begin{subequations}\label{con:expected_precise}
 \begin{equation}\label{con:pignistic_precise}
  \mathbb{E}_m(f_{\omega_i})=\sum_{j=1}^M u_{ij}BetP_m(\{\omega_j\})
 \end{equation}
 where  $BetP_m$ is the \emph{pignistic} probability measure computed from mass function $m$ by   the \emph{pignistic  transformation}, defined as
 \begin{equation}\label{con:probability_transformation}
  BetP_m(\{\omega_j\})=\sum_{\{A \subseteq \Omega : \omega_j \in A\}} \frac{m(A)}{|A|},
 \end{equation}
 for all $\omega_j\in \Omega$. Other decision criteria in the belief function framework are reviewed in \cite{denoeux2019decision} and \cite{MA2021106742}.
\end{subequations}

\subsection{Evidential neural network}
\label{sec:DS_layer}

Den{\oe}ux \cite{denoeux2000neural} proposed a distance-based neural-network based on DS theory, known as the \emph{evidential neural network (ENN) classifier}. The ENN classifier summarizes the learning set by a small number of prototypes, and treats the proximity of an input vector to each prototype as a piece of evidence about its class. The different pieces of evidence are represented by mass functions, which are combined using Dempster's rule \eqref{con:dempster}. This section provides a brief description of the ENN classifier.

We consider a training set ${{\cal X}=\left\{\boldsymbol{x}^1,\ldots,\boldsymbol{x}^N\right\}\subset\mathbb{R}^P}$  of $N$ examples represented by $P$-dimensional feature vectors, and $n$ prototypes $\{\boldsymbol p^1,\ldots,\boldsymbol p^n\} \subset\mathbb{R}^P$. 
For a test sample $\boldsymbol x$, the ENN classifier constructs mass functions that quantify the uncertainty about its class in  $\Omega=\{\omega_1,\ldots,\omega_M\}$, using a three-step procedure. This procedure can  be implemented in a neural-network layer, which will be plugged into an FCN model as explained in Section \ref{sec:network_architecture}. The three-step procedure can be described as follows.

\begin{description}
\item[Step 1:] The similarity between $\boldsymbol x$ and each prototype $\boldsymbol p^l$ is computed as
\begin{equation}\label{con:si}
 s^l=\alpha^l\exp\left(-\left(\eta^ld^l\right)^2\right), \quad l=1,\ldots,n,
\end{equation}
where $d^l={\left\|\boldsymbol x-\boldsymbol p^l\right\|}$ is the Euclidean distance between $\bx$ and prototype $\bp^l$, $\eta^l \in \mathbb{R}$ is a scale parameter and  $\alpha^l$ is a parameter in  $[0,1]$. Prototypes $\boldsymbol p^1,\ldots,\boldsymbol p^n$ can be considered as vectors of connection weights between the input layer and a hidden layer of $n$  Radial Basis Function (RBF) units. The number $n$ of prototypes is a hyper-parameter and can be tuned using a validation set or by cross-validation.
\item[Step 2:] The mass function $m^l$ associated to reference pattern $\boldsymbol p^l$ is computed as
\begin{subequations}
 \label{con:m^i}
 \begin{align}
  m^{l}(\{\omega_j\})& = v_j^l s^{l}, \quad j=1,\ldots,M\\
  m^{l}(\Omega)& =1-s^{l},
 \end{align}
\end{subequations}
where $v_j^l\ge 0$ is the degree of membership of prototype $\bp^l$ to class $\omega_j$ with $\sum_{j=1}^Mv_j^l=1$.
We denote the vector of masses induced by prototype $\bp^l$ as
\[
\boldsymbol m^{l}=(m^{l}(\{\omega_1\}), \ldots,m^{l}(\{{\omega}_{M}\}),m^{\mathit l}(\Omega))^T.
\]
Eq. \eqref{con:m^i} can be regarded as computing the activation of units in a ``mass functions'' layer  composed of $n$ modules of $M+1$ units each. The activations of the units in module $l$ correspond to the belief masses assigned by $m^l$.
\item[Step 3:] The $n$ mass functions $\boldsymbol m^l$, $l=1,\ldots,n$, are aggregated by Dempster's rule \eqref{con:dempster}. The combined mass function can be computed iteratively as $\mu^1=m^1$ and $\mu^l=\mu^{l-1}\cap m^l$ for $l=2,\ldots,n$. From \eqref{con:dempster1}, we have
\begin{subequations}
 \label {con:BFoutput}
 \begin{multline}
  \mu^l(\{\omega_j\})=\mu^{l-1}(\{\omega_j\})m^{ l}(\{\omega_j\})+\\
  \mu^{l-1}(\{\omega_j\})m^{l}(\{\Omega\})+\mu^{l-1}(\Omega)m^{ l}(\{\omega_j\})        \label{con:mu^ij}
 \end{multline}
 for  $l=2,\ldots,n$ and $j=1,\ldots,M$, and
 \begin{equation}
  \mu^l(\Omega)=\mu^{l-1}(\Omega)m^{l}(\Omega) \quad l=2,\ldots,n.
  \label{con:mu^iM}
 \end{equation}
\end{subequations}
The output vector $\boldsymbol m=(m(\{\omega_1\}), \ldots,m(\{{\omega}_{ M}\}),m(\Omega))^T$   is finally obtained by normalizing $\mu^n$ as
\[
m(A)=\frac{\mu^n(A)}{\mu^n(\Omega) + \sum_{j'=1}^M \mu^n(\{\omega_{j'}\})},
\]
with $A \in \{ \{\omega_1\},\ldots,\{\omega_M\}, \Omega \}$.
\end{description}

The network parameters are the prototypes $\boldsymbol p^l$, the coefficients $\alpha^l$ and $\eta_l$, and the membership degrees $v^l_j$ for $l=1,\ldots,n$ and $j=1,\ldots,M$. They are learnt by minimizing a loss function. To enforce the constraints $0\le \alpha^l\le1$, we introduce new variables $\xi^l \in \mathbb{R}$ such that
\[
\alpha^l=\frac{1}{1+\exp(-\xi^l)} \in (0,1).
\]
Similarly, the constraints on parameters $v^l_j$ are enforced by  introducing new parameters $\delta_j^l \in \mathbb{R}$ such that
\begin{equation}\label{con:vjl}
v_j^l=\frac{(\delta_j^l)^2}{\sum_{j'=1}^M (\delta_{j'}^l)^2}
\end{equation}
for $l=1,\ldots,n$ and $j=1,\ldots,M$. More details can be found in \cite{denoeux2000neural}.

\subsection{Fully convolutional network}
\label{sec:fcn}

The performance of an ENN classifier in semantic segmentation tasks heavily depends on the information contained in its input features. Feature representation, an essential part of the machine learning workflow, consists in discovering the predictors needed for semantic segmentation from input images. In recent years, FCNs \cite{long2015fully} and their variants \cite{noh2015learning,krahenbuhl2011efficient} have achieved remarkable performances thanks to  their ability to construct rich pixel-wise deep feature representations.

FCNs owe their name to their architecture, which is built only from locally connected layers, such as convolution, pooling, and upsampling layers. No dense layer is used in this kind of architecture. Generally, an FCN consists of two main parts: an encoder-decoder architecture for pixel-wise object representation and a softmax layer for pixel-wise assignments. In the encoder-decoder architecture, an input image is encoded by several convolutional and pooling layers and then decoded by one or more upsampling layers. The softmax layer assigns each pixel in the input image to one of the classes based on the outputs of the encoder-decoder architecture. Therefore, the outputs of the encoder-decoder architecture, called the \emph{pixel-wise feature maps}, are considered as a feature representation of the input image. In the study, these feature maps are used as input to a DS layer allowing for set-valued semantic segmentation, as will be shown in Section \ref{sec:network_architecture}.

\begin{figure}
	\centering
	\includegraphics[width=0.9\linewidth]{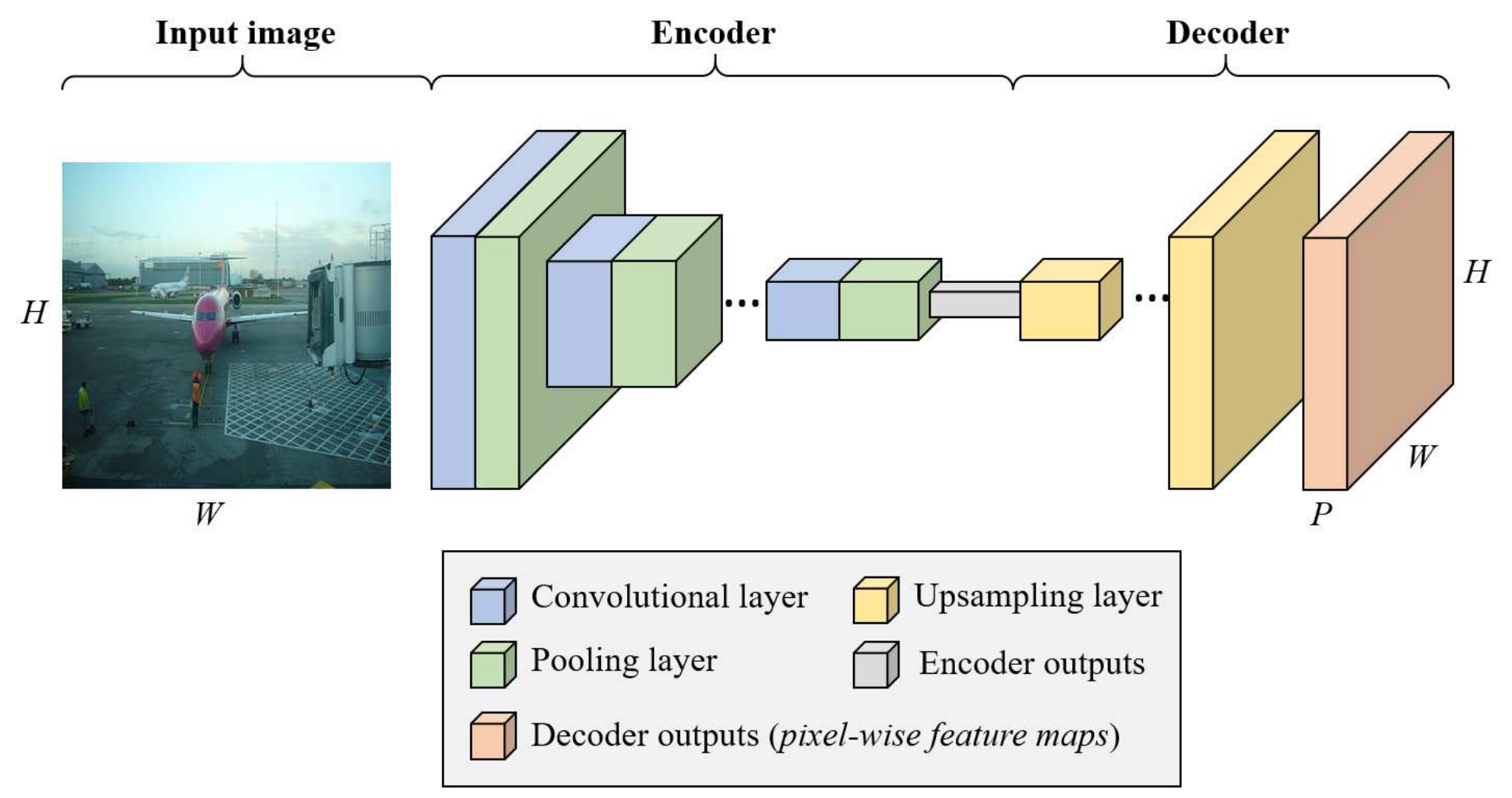}
	\caption{An illustration of the encoder-decoder architecture. An encoder downsizes its input by convolution and pooling operations. The outputs of the encoder, as the sparse feature maps, are imported into a decoder. A decoder upsamples and densifies its inputs by performing the reverse operation of convolution and pooling. The final decoder outputs are the pixel-wise feature maps.}\label{fig:encoder_decoder}
\end{figure}

To understand the feature representation of FCNs, we briefly recall the encoder-decoder architecture illustrated in Figure \ref{fig:encoder_decoder}. The encoder part consists of several convolutional and pooling layers. Each convolutional layer performs convolutions its input to produce a set of feature maps. Let $\boldsymbol z=(z^1,\dots,z^D)$ be the input made up of $D$ input maps or \emph{input channels} $z^i$ ($i=1,\dots,D$) of size $H \times W$. The processes in a convolutional layer with input $\boldsymbol z$, consisting of $e$ convolution kernels with size $a \times b$, are expressed as
\begin{equation}\label{con:convolution}
c^j=f(\lambda^j+\sum_i\varepsilon^{i,j}\ast z^i),
\end{equation}
where $\varepsilon^{i,j}$, a  matrix of size $a\times b$, is the convolution kernel between the $i$-th input map and the $j$-th output map; $\lambda^j$ is the bias of kernel $\varepsilon^{i,j}$; $\ast$ denotes the convolution operation;  $c^j$ is the $j$-th output feature map, with size  $\frac{h-a+1}{r} \times \frac{w-b+1}{r}$, $j=1,\dots,e$; $r$ is the stride with which the kernel slides over input map $z^i$, and $f$ is the activation function, such as the rectified linear unit $\textsf{ReLU}(x)=\max(0,x)$ \cite{5459250}. A pooling layer  follows the convolutional layer to sub-sample feature map $c^j$ by computing some statistics of feature values within non-overlapping $s \times s$  windows. In the case of max-pooling used in this paper, the statistic is the maximum.  Thus, the outputs of the pooling layer is composed of the $D$ feature maps sub-sampled by factor $s$. For example, feature map $c^j$ with size $\frac{h-a+1}{r} \times \frac{w-b+1}{r}$ is downsized to $\frac{h-a+1}{2r} \times \frac{w-b+1}{2r}$ by a pooling layer with a $2 \times 2$ non-overlapping window.

Although the convolution and pooling operations in the encoder part help feature representation by retaining only robust activations, spatial information within a receptive field is lost, which may be critical for image semantic segmentation. To address the issue, a decoder part made up of one or more upsampling layers is added at the output of the encoder part. The decoder performs the reverse operation of convolution and pooling for reconstructing a set of activations with the same size of the input image, as shown in Figure \ref{fig:encoder_decoder}. Thus, the outputs of the decoder part are enlarged feature maps. In the study, we use a \emph{deconvolution layer}  \cite{noh2015learning} to implement the upsampling operation.

A deconvolutional layer densifies its inputs of sparse feature maps through convolution-like operations with multiple learned kernels. However, contrary to convolutional layers, which connect multiple inputs within a kernel to a single activation, a deconvolutional layer associates a single input in a feature map to multiple outputs. Thus, the outputs of a deconvolutional layer are enlarged and dense feature maps. The processes of a deconvolution operation can also be summarized as Eq. \eqref{con:convolution}, but its kernel sizes are larger than the input sizes, i.e.,  $a\geq H$ and $b\geq W$.

\section{Evidential fully convolutional network}
\label{sec:efcn}

In this section, we describe the proposed E-FCN. Section \ref{sec:network_architecture} presents the overall architecture composed of an encoder-decoder module for feature representation, a DS layer to construct mass functions, and a utility layer for decision-making. The details of the utility layer are described in  Section \ref{sec:Ulayer}. Section \ref{sec:learning} introduces the strategy for training E-FCN models using a learning set with soft labels.

\subsection{Network architecture}
\label{sec:network_architecture}

\begin{figure}
 \centering
 \includegraphics[width=\linewidth]{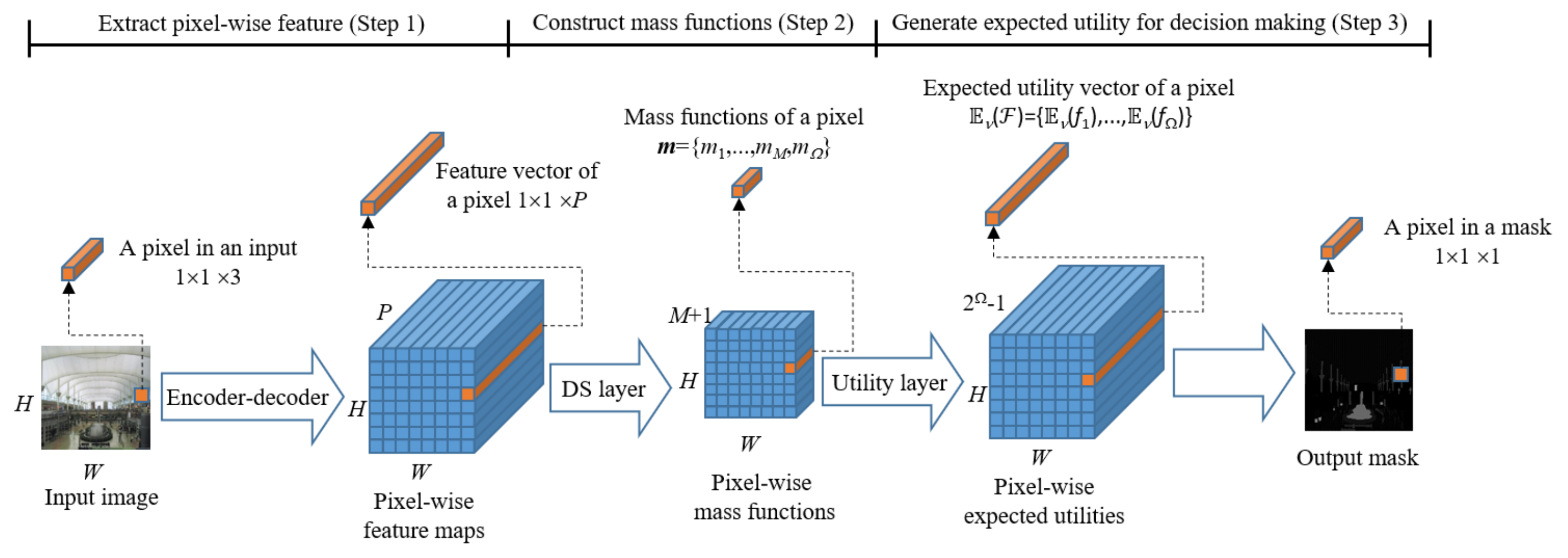}
 \caption{Architecture of an evidential fully convolutional network (E-FCN). The E-FCN performs semantic segmentation using a three-step procedure. In the first step, an encoder-decoder architecture extracts pixel-wise feature maps from the input image. Each vector in the feature maps is fed into a DS layer to construct the pixel-wise mass functions in the second step. These mass functions are finally fed into a utility layer to generate the pixel-wise expected utilities of all acts. Finally, the segmentation mask is computed based on the expected utilities.}\label{fig:evidential_FCN}
\end{figure}

The main idea of this work is to hybridize the ENN classifier presented in Section \ref{sec:DS_layer} and the FCN recalled in Section \ref{sec:fcn} by ``plugging'' a DS layer followed by a utility layer at the output of the final deconvolutional layer in the FCN. The architecture of the proposed method, called the \emph{evidential FCN} (E-FCN), is illustrated in Figure \ref{fig:evidential_FCN}. An E-FCN classifier performs set-valued semantic segmentation and quantifies the uncertainty about the class of each pixel, taking values in $\Omega=\{\omega_1,\ldots,\omega_M\}$, using a three-step procedure defined as follows.

\begin{description}
 \item{Step 1:} As in a probabilistic FCN (P-FCN), an image of size $W \times H  \times 3$ is  presented as input to the the encoder-decoder architecture of an FCN to generate pixel-wise feature maps of size  $W \times H  \times P$, where $P$ is the number of \emph{output channels}. Each feature vector $1 \times 1 \times P$ from a pixel-wise feature map is a $P$-dimensional representation of the corresponding pixel, ready to be fed into the DS layer. This architecture generates reliable pixel-wise representations of the input image. Thanks to the representations, the E-FCN yields similar or even better performance for precise semantic segmentation than does a P-FCN with the same encoder-decoder architecture, as will be shown in Section \ref{sec:precise_segmentation}.
 
 \item{Step 2:}  Each feature vector from the encoder-decoder architecture is fed into the DS layer, in which it is converted into  a mass function as explained in Section \ref{sec:DS_layer}. The output of the DS layer for a given feature vector is an $(M+1)$-dimensional mass vector
 \[
 \boldsymbol m=(m(\{\omega_1\}), \ldots,m(\{{\omega}_{ M}\}),m(\Omega))^T.
 \]
 Thus, given pixel-wise feature maps of size $W \times H  \times P$ from Step 1, the output of the DS layer is a tensor of size $W\times H\times (M+1)$. Each mass vector in the tensor represents the uncertainty about the class of the corresponding pixel. More precisely, the mass $m(\{\omega_i\})$ is a degree of belief that the ground truth of the pixel is $\omega_i$. The DS layer tends to allocate uniform masses if the representations contain confusing information. The additional degree of freedom $m(\Omega)$ makes it possible to quantify the lack of evidence \cite{denoeux2019logistic} and verify whether the model is well trained \cite{tong2019ConvNet}. The advantages of this uncertainty representation will be demonstrated in the performance evaluation of set-valued semantic segmentation using E-FCN in Section \ref{sec:imprecise_segmentation}.
 
 \item{Step 3:} The output pixel-wise mass vectors are fed into a utility layer for semantic segmentation, which is used to compute the expected utility of acts. Each act is defined as the assignment of a pixel to a non-empty subset $A$ of $\Omega$. Therefore, the output of the layer for each feature vector from Step 2 is an expected-utility vector at most equal to $2^\Omega-1$ when all of the possible acts are considered. The utility layer allows the E-FCN to perform set-valued semantic segmentation. This capability will be demonstrated by the performance comparison between the two types of FCNs in the tasks of set-valued segmentation (Section \ref{sec:imprecise_segmentation}) and novelty detection (Section \ref{sec:novelty_detection}). More details of the utility layer for set-valued segmentation are introduced in the next section.
\end{description}

\subsection{Utility layer for decision making}
\label{sec:Ulayer}

In this section, we describe in greater detail the decision-making process taking place in the utility layer. Section \ref{sec:Ulayer} begins with introducing the precise semantic segmentation method using mass functions and utilities. Section \ref{sec:Ulayer} describes a method for computing the utility of set-valued pixel-wise classification, after which an approach to set-valued classification based on mass functions is described in Section \ref{sec:Ulayer}. In Section \ref{sec:Ulayer}, we summarize the work
ow as a neural network layer for the E-FCN model.


\subsubsection{Precise semantic segmentation}
\label{sec:precise_classification}

Let $\Omega=\{\omega_1,\dots,\omega_M\}$ be the set of classes. For semantic segmentation problems with  precise prediction, each pixel in an image is assigned to exactly one  class. An act is thus defined as the assignment of a pixel to one and only one of the $M$ classes, and the set of acts is $\mathcal{F}=\{f_{\omega_1},\dots,f_{\omega_M}\}$, where $f_{\omega_i}$ denotes assignment to class $\omega_i$. To make decisions, we define a utility matrix $\bU$ of size $M \times M$, whose general term $u_{ij} \in [0,1]$ is the utility of assigning a pixel to class $\omega_i$ when the true class is $\omega_j$. 

When uncertainty about $\Omega$ is described by belief functions, each act $f_{\omega_i}$ induces expected utilities, such as the pignistic expected utilities defined by Eq. \eqref{con:expected_precise}. Given utility matrix $\bU$  and the output of the DS layer $\boldsymbol m$ for a given pixel, the pignistic expected utility of assigning that pixel to class $\omega_i$ is
\begin{equation}\label{con:expected_pignistic_precise}
\mathbb{E}_m(f_{\omega_i})=\sum_{j=1}^M u_{i,j}BetP_m(\{\omega_j\}),
\end{equation}
where $BetP_m$ is the pignistic probability defined by Eq. \eqref{con:probability_transformation}. The pixel is finally assigned to set class $\omega_i$ such that
\begin{equation}
\label{eq:i}
i^{\ast}=\arg\max_{\omega_i \in \{1,\dots,M\}}\mathbb{E}_m(f_{\omega_i}).
\end{equation}

\subsubsection{Extending the utility matrix}
\label{sec:owa}

For semantic segmentation problems with imprecise prediction, we adopt the approach described in \cite{MA2021106742} for set-valued classification under uncertainty, which allows the assignment of a pixel to any non-empty subset $A$ of $\Omega$. The set of acts thus potentially becomes $\mathcal{F}=\{f_A : A \subseteq \Omega, A\neq  \emptyset\}$, where $f_A$ denotes the assignment to a subset $A$. (In practice, when the cardinality of $\Omega$ is very large, we may only consider acts $f_A$ for \emph{some} subsets $A$ of $\Omega$). In this study, $f_A$ is referred to as an \emph{imprecise assignment} when subset $A$ is a \emph{multi-class set} with $|A| \geq 2$. For decision-making with $\mathcal{F}$, the utility matrix $\bU$  has to be extended to a matrix $\widetilde{\bU}$ of size $(2^M-1) \times M$, where each element $\widetilde{u}_{A,j}$ denotes the utility of assigning a pixel to set $A$ of classes when the true label is $\omega_j$. Following \cite{MA2021106742}, this extension is performed as follows.

When the true class is $\omega_j$, the utility of assigning a pixel to set $A$ is defined as an Ordered Weighted Average (OWA) aggregation \cite{yager1988ordered} of the utilities of each precise assignment in $A$ as
\begin{equation}\label{con:owa}
{\widetilde u}_{A,j}=\sum_{k=1}^{\left|A\right|}g_k \, u_{(k)j}^A,
\end{equation}
where $u_{(k)j}^A$ is the $k$-th largest element in the set $\{u_{ij} : \omega_i \in A\}$ made up of the elements in the utility matrix $\bU$, and weights $\boldsymbol g=(g_1,\ldots,g_{|A|})$ represent the preference  to choose $u_{(k)j}(A)$ if forced to select
 a single value in $\{u_{ij} : \omega_i \in A\}$. The components of weight vector $\boldsymbol g$ represent the \emph{tolerance to imprecision} of a decision maker (DM). For example, full tolerance to imprecision is achieved when the assignment act $f_A$ has utility 1 once set $A$ contains the true label, no matter how large $A$ is. In this case, only the maximum utility of elements in set $\{u_{ij},\omega_i \in A\}$ is considered: $(g_1,g_2,\dots,g_{|A|})=(1,0,\dots,0)$. At the other extreme,  a DM attaching no value to imprecision would consider the act $f_A$ as equivalent to selecting one class uniformly at random from $A$: this is achieved when
\[
(g_1,g_2,\dots,g_{|A|})=\left(\frac1{|A|},\frac1{|A|},\dots,\frac1{|A|}\right),
\]
in which case the OWA operator becomes the average. In this study, following \cite{MA2021106742}, we determine the weight vector $\boldsymbol g$ of the OWA operator by adapting O'Hagan's method \cite{ohagan88}. We define the  tolerance to imprecision as
\begin{equation}\label{TDI}
TDI(\boldsymbol g)=\sum_{k=1}^{\left|A\right|}\frac{\left|A\right|-k}{\left|A\right|-1}g_k=\gamma,
\end{equation}
which equals  1 for the maximum, 0 for the minimum, and 0.5 for the average. In practice, we only need to consider values of $\gamma$ between 0.5 and 1 as a precise assignment is always more desirable than an imprecise one when $\gamma \textless 0.5$ \cite{MA2021106742}. Given a value of $\gamma$, we can compute the weights of the OWA operator by maximizing the entropy
\begin{equation}\label{ENT}
ENT(\boldsymbol {g})=-\sum_{k=1}^{\left|A\right|} g_k  \log g_k
\end{equation}
subject to the constraints $TDI(\boldsymbol {g})=\gamma$, $\sum_{k=1}^{\left|A\right|}g_k=1$, and $g_k \geq 0$.

\begin{Ex}\label{ex:nosoftlable}
 Table \ref{tab:example_utility} shows an example of the extended utility matrix generated by an OWA operator with $\gamma=0.8$. The first three rows constitute the original utility matrix, indicating that the utility equals 1 when assigning a sample to its true class, and 0 otherwise. The remaining rows are the matrix of the aggregated utilities. For example, we get a utility of 0.8  when assigning a sample to set $\{\omega_1,\omega_2\}$ if the true label is $\omega_1$.
 
 \begin{table}
  \centering
  \caption{Utility matrix extended by an OWA operator with $\gamma=0.8$.}\label{tab:example_utility}
  \begin{tabular*}{\hsize}{@{}@{\extracolsep{\fill}}cccc@{}}
   \hline
   \multirow{2}{*}{}           & \multicolumn{3}{c}{Classes} \\ \cline{2-4}
   & $\omega_1$ & $\omega_2$ & $\omega_3$ \\ \hline
   $f_{\{\omega_1\}}$          & 1          & 0          & 0          \\
   $f_{\{\omega_2\}}$          & 0          & 1          & 0          \\
   $f_{\{\omega_3\}}$          & 0          & 0          & 1          \\
   $f_{\{\omega_1,\omega_2\}}$ & 0.8        & 0.8        & 0          \\
   $f_{\{\omega_1,\omega_3\}}$ & 0.8        & 0          & 0.8         \\
   $f_{\{\omega_2,\omega_3\}}$ & 0          & 0.8        & 0.8        \\
   $f_\Omega$            & 0.6819     & 0.6819     & 0.6819     \\ \hline
  \end{tabular*}
 \end{table}
\end{Ex}

\subsubsection{Set-valued semantic segmentation using belief function and utility theory}
\label{sec:imprecise_classification}

Based on an extended utility matrix $\btU$ and the output of the DS layer $\boldsymbol m$ for a given pixel, we can compute the pignistic expected utility of  assigning that pixel to set $A$   as
\begin{equation}\label{con:expected_imprecise}
 \mathbb{E}_m(f_{A})=\sum_{j=1}^M \widetilde{u}_{A,j}BetP_m(\{\omega_j\}),
\end{equation}
where $BetP_m$ is the pignistic probability defined by Eq. \eqref{con:probability_transformation}. The pixel is finally assigned to set $A$ such that 
\begin{equation}
\label{eq:A}
A=\arg\max_{\emptyset\neq B\subseteq\Omega}\mathbb{E}_m(f_{B}).
\end{equation}

\subsubsection{Utility layer}
\label{sec:utility_layer}

The procedure of assigning a pixel to a set of classes using utility theory is implemented as a layer of the neural network, called a \emph{utility layer}. In this layer, the inputs and outputs are, respectively, the pixel-wise mass vectors $\boldsymbol m$ from the preceding DS layer and the pixel-wise expected utilities of all acts in $\mathcal{F}$. The connection weight between unit $j$ of the DS layer and output unit $A\subseteq \Omega$ corresponding to the assignment to set $A$ is the utility value ${\widetilde u}_{A,j}$. As coefficient $\gamma$ describing the imprecision tolerance degree is fixed, the connection weights of the expected utility layer are fixed and do not need to be updated during training. 

In practice, the connections between the DS and utility layers  can be determined by the user. For example, one can build a utility layer using the utility values ${\widetilde u}_{A,j}$ with $|A|=1$ to only consider precise assignments, or $0<|A|\le 2$ to consider assignment to sets classes of cardinality one or two. In this paper, we have only considered  the acts $f_A$ such that $A$ is a singleton, $\Omega$, or one of the soft labels present in the learning set (as explained in Section \ref{sec:learning} below).

\subsection{Learning with soft labels}
\label{sec:learning}

In traditional learning systems for image semantic segmentation, all pixels are labeled with a single class even when their true class cannot be determined with full certainty. For example,  the true class may be uncertain at object borders, but the border pixels are still given precise labels. Additionally, one cannot reliably label  some small objects in an image, such as distant objects in a driving scene. Arbitrarily giving precise labels to  pixels with confusing information may have negative effects on learning systems for image semantic segmentation. The notion of \emph{soft label} \cite{come09,denoeux19f} may be a way to solve this problem.

Here, we define a soft label as a nonempty subset  $A_\ast \in 2^\Omega \backslash \emptyset$ of classes a pixel may belong to, based on our current knowledge. For example, label $A_\ast=\{\omega_i,\omega_j\}$ indicates that the true class of a pixel is known to be either $\omega_i$ or $\omega_j$ but we cannot determine which one specifically. A strategy of end-to-end learning is proposed to train an E-FNC from an image learning set with soft labels. All parameters in the DS layer are  first initialized randomly using normal distributions.  For a given pixel with nonempty soft label  $A_\ast \subseteq \Omega $,  let $m_l$ be the logical mass function with focal set $A_\ast$, i.e., such that $m_l(A_\ast)=1$. The \emph{labeling} pignistic expected utilities $\mathbb{E}_{\boldsymbol{m}_l}(f_{A})$ for  $A \in 2^ \Omega\backslash {\emptyset}$ can be computed using Eq. \eqref{con:expected_imprecise} and the pignistic belief-probability transformation Eq. \eqref{con:probability_transformation}. Similarly, we consider the \emph{predicted} pignistic expected utilities $\mathbb{E}_{m}(f_{A})$ for $A \in 2^ \Omega\backslash \emptyset$ , where $m$ is the predicted mass function from the DS layer of the E-FCN, with focal sets $\{\omega_1\},\ldots,\{\omega_M\},\Omega$. For a given pixel with soft label $m_l$ and predicted mass function $m$, the loss $\mathcal{L}(m,m_l)$  is defined as the squared Euclidean distance between the vectors of expected utilities w.r.t. $m_l$ and $m$:
\begin{equation}
\mathcal{L}(m,m_l)=\sum_{\emptyset\neq A \subseteq \Omega}\left[\mathbb{E}_{m_l}(f_{A})-\mathbb{E}_{m}(f_{A})\right]^2.
\end{equation}

The derivatives of $\mathcal{L}_p(m,m_l)$ of the error w.r.t the output masses  $m(\{\omega_k\})$ are 
 \begin{equation}
 \label{con:derivative_eulayer}
 \begin{aligned}
 \frac{\partial \mathcal{L}(m,m_l)}{\partial m(\{\omega_k\})}&=\sum_{\emptyset\neq A \subseteq \Omega} \frac{\partial \mathcal{L}(m,m_l)}{\partial \mathbb{E}_{m}(f_{A})}\cdot \frac{\partial \mathbb{E}_{m}(f_{A})}{\partial m(\{\omega_k\})}\\
 &=-2\sum_{\emptyset\neq A \subseteq \Omega} \left[\mathbb{E}_{m_l}(f_{A})-\mathbb{E}_{m}(f_{A})\right] \sum_{j=1}^M \deriv{\mathbb{E}_{m}(f_{A})}{BetP_m(\omega_j)} \deriv{BetP_m(\omega_j)}{m(\{\omega_k\})}\\
 &=-2\sum_{\emptyset\neq A \subseteq \Omega} \left[\mathbb{E}_{m_l}(f_{A})-\mathbb{E}_{m}(f_{A})\right] \sum_{j=1}^M \widetilde{u}_{A,j}\left(\delta_{kj}-\frac1M\right),
 \end{aligned}
 \end{equation}
 where $\delta_{kj}=1$ if $k=j$ and $\delta_{kj}=0$ otherwise. 
The derivatives of $m(\{\omega_k\})$ w.r.t $p^l_k$, $\eta^l$, and $\xi^l$ in the DS layer are the same as in Den{\oe}ux's original work   \cite{denoeux2000neural}, and the gradient with respect to all network parameters can be back-propagated from the output layer to the input layer. 
%

\section{Experiments}
\label{sec:experiments}

In this section, we present  numerical experiments that demonstrate the advantages of the proposed model. The databases and metrics are first  introduced in Section \ref{sec:databases_metrics}. Precise and imprecise segmentation results are then reported, respectively, in Sections \ref{sec:precise_segmentation} and \ref{sec:imprecise_segmentation}. Finally,  novelty detection results are presented in Section \ref{sec:novelty_detection}.

\subsection{Databases and metrics for performance evaluation}
\label{sec:databases_metrics}

\subsubsection*{Databases}
\label{sec:databases}

 Three benchmark databases were used in the study: Pascal VOC 2011 \cite{everingham2015pascal}, MIT-scene Parsing \cite{zhou2016semantic}, and SIFT Flow \cite{tighe2010superparsing}. These databases were used to train and test the E-FCNs as well as  probabilistic FCNs (P-FCNs) for comparison.

The Pascal VOC 2011 database contains 20 object classes in 5034 images, with   segmentation masks that indicate the  class of each pixel, or label it as ``background''  if the object does not belong to one of the twenty specified classes. The MIT-scene Parsing and SIFT Flow databases are similar to the Pascal VOC 2011 database but have, respectively,   150 categories in 20K images   and 33 classes in 2688 images.  The list of classes for the three databases are given in Table \ref{tab:class_lists}. Each of the three databases was split into 50\% for training/validation and 50\% for testing. In the study, the validation sets were used to determine hyper-parameters, such as the number of prototypes in each DS layer. In practice, a validation set can also be used to determine the optimal tolerance to imprecision $\gamma$ since it can also be considered as a hyper-parameter.

\begin{table}[]
\caption{Lists of classes for the three databases used in this study. Classes in bold characters are  included in two or three databases. Classes with close meanings, such as  ``minibike'' and ``motorbike'', are considered as identical.}\label{tab:class_lists}
	\begin{tabular}{lp{0.75\textwidth}ll}
	\cline{1-2}
		Database          & Class list                                                                                                                                                                                                                                                                                                                                                                                                                                                                                                                                                                                                                                                                                                                                                                                                                                                                                                                                                                                                                                                                                                                                                                                                    &  \\  \cline{1-2}  
		Pascal VOC 2011   & background, cat, dog, horse, sheep, train, \textbf{sofa}, \textbf{aeroplane}, \textbf{bicycle}, \textbf{bird}, \textbf{boat}, \textbf{bottle}, \textbf{bus}, \textbf{car}, \textbf{chair}, \textbf{cow}, \textbf{diningtable}, \textbf{motorbike}, \textbf{person}, \textbf{pottedplant}, \textbf{tv}. &  \\ \cline{1-2} 
		MIT-scene parsing & wall,  floor, ceiling, bed, cabinet, earth, curtain, water, painting, shelf, house, mirror, rug, armchair, seat, desk,  wardrobe, lamp, bathtub, railing, cushion, base, box, column, chest, counter, sink, skyscraper, fireplace, refrigerator, grandstand, path, stairs, runway, case, pool, pillow, screen, bookcase, blind, coffee, toilet, flower, book, hill, bench, countertop, stove, palm, kitchen, computer, swivel, bar, arcade, hovel, towel, light, truck, tower, chandelier, booth, dirt track, apparel, land, bannister, escalator, ottoman, buffet, poster, stage, van, ship, fountain, conveyer, canopy, washer, plaything, swimming, stool, barrel, basket, waterfall, tent, bag, minibike, cradle, oven, ball, food, step, tank, trade, microwave, pot, animal, lake, dishwasher, screen, blanket, sculpture, hood, sconce, vase, traffic, tray, ashcan, fan, pier, screen, plate, monitor, bulletin, shower, radiator, glass, clock, flag, \textbf{sofa}, \textbf{airplane}, \textbf{building}, \textbf{sky}, \textbf{tree}, \textbf{road}, \textbf{windowpane}, \textbf{grass}, \textbf{sidewalk}, \textbf{person}, \textbf{door}, \textbf{table}, \textbf{mountain}, \textbf{plant}, \textbf{chair}, \textbf{car}, \textbf{sea}, \textbf{field}, \textbf{fence}, \textbf{rock}, \textbf{sign}, \textbf{sand}, \textbf{staircase}, \textbf{river}, \textbf{bridge}, \textbf{boat}, \textbf{bus}, \textbf{awning}, \textbf{streetlight}, \textbf{tv}, \textbf{pole}, \textbf{bottle}, \textbf{minibike}, \textbf{bicycle}. &  \\ \cline{1-2} 
		SIFT Flow         & balcony, crosswalk, desert, moon, sun,  window, \textbf{awning}, \textbf{bird}, \textbf{boat}, \textbf{bridge}, \textbf{building}, \textbf{bus}, \textbf{car}, \textbf{cow}, \textbf{door}, \textbf{fence}, \textbf{field}, \textbf{grass}, \textbf{mountain}, \textbf{person}, \textbf{plant}, \textbf{pole}, \textbf{river}, \textbf{road}, \textbf{rock}, \textbf{sand}, \textbf{sea}, \textbf{sidewalk}, \textbf{sign}, \textbf{sky}, \textbf{staircase}, \textbf{streetlight},  \textbf{tree}. & \\ \cline{1-2} 
	\end{tabular}
\end{table}

\begin{figure}
 \centering
 \subfloat[\label{fig:softlabel_voc}]{\includegraphics[width=\textwidth]{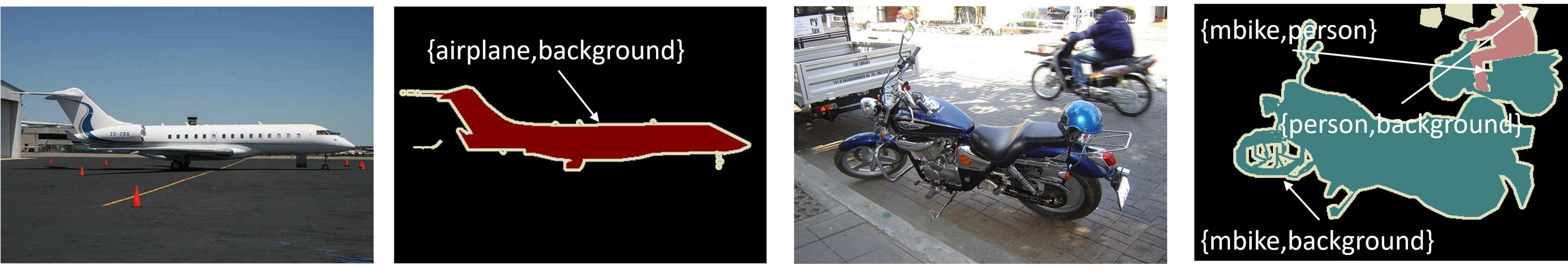}}\\
 \subfloat[\label{fig:softlabel_mit}]{\includegraphics[width=0.49\textwidth]{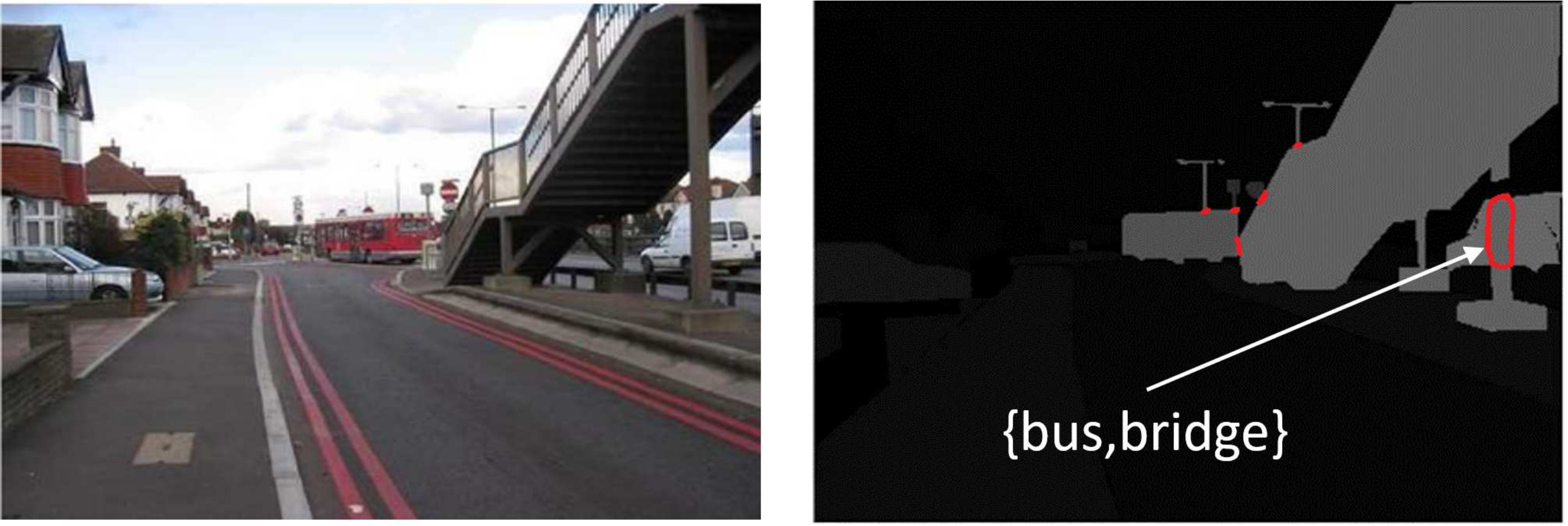}}\hspace{0.01\textwidth}
 \subfloat[\label{fig:softlabel_sift}]{\includegraphics[width=0.49\textwidth]{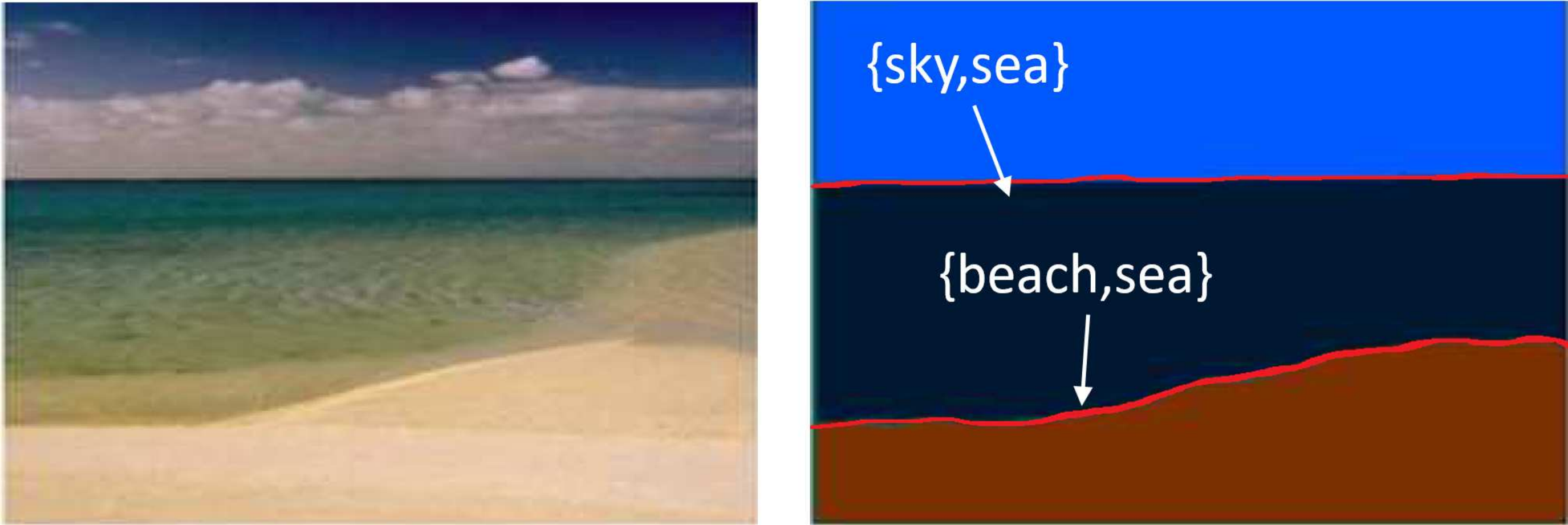}}
 \caption{Segmentation masks with soft labels: (a) Pascal VOC 2011, (b) MIT-scene Parsing, and (c) SIFT Flow.}\label{fig:soft_label}
\end{figure}

There is no confidence value associated with the pixel labels in any of the three databases. Thus, we defined  soft labels for them. For the Pascal VOC 2011 database, we assigned each pixel in a boundary area a soft label $A \subseteq \Omega$, where $A$ consists of the object classes around the boundary area. Some examples are shown in Figure \ref{fig:softlabel_voc}. For the MIT-scene Parsing and SIFT Flow databases with no identified boundary areas, we assigned soft labels to the pixels situated between every two objects, as shown in Figures \ref{fig:softlabel_mit} and \ref{fig:softlabel_sift}.

A semantic segmentation model should not only be accurate for the classes in the learning set,  but it  should also be able to detect some objects whose classes are not included in the learning set. To evaluate this novelty detection capacity,  we mixed the three databases: for example, an FCN model trained using the Pascal VOC 2011 database was tested on the other two databases. 

\subsubsection*{Metrics}
\label{sec:metrics}

We used three metrics for the performance evaluation of semantic segmentation: pixel utility (PU), utility of intersection over union (UIoU), and expected calibration error (ECE).

\paragraph{Pixel utility.} For an image with $T$ pixels, the \emph{pixel utility}  is defined as
\begin{equation}\label{con:pu}
PU=\frac1{|T|}\sum_{i=1}^{|T|}{\widetilde u}_{A(i),A_\ast(i)}
\end{equation}
where $A_\ast(i)$ is the label of pixel $i$, $A(i)$ is the selected set of classes for pixel $i$ determined from Eq. \eqref{eq:A}, and using the notations introduced in Section \ref{sec:Ulayer}, ${\widetilde u}_{A(i),A_\ast(i)}$ is the utility of assigning pixel $i$ to subset $A(i) \subseteq \Omega$ when its label is $A_\ast(i)$. Thus, PU is the same as pixel accuracy when only considering precise assignments and precise labels.  To consider soft labels, the utility matrix $\btU$ defined in Section \ref{sec:Ulayer} should be extended to a matrix  $\btU'$ of size $(2^M-1) \times (2^M-1)$ with general term $\widetilde{u}_{A,A_\ast}$ defined as  the utility of assigning a pixel to subset $A \subseteq \Omega$ when its label is $A_\ast$, with $|A_\ast| \geq 1$. Soft label $A_\ast$ means that we only know the true class of a pixel is in set $A_\ast$, and nothing more. To define the utility $\widetilde{u}_{A,A_\ast}$, we first compute the average the average of the utilities of selecting subset $A$ when the true class is in $A_{\ast}$ as
\begin{subequations}
\begin{equation}
\overline{u}_{A,A_\ast}=\frac{1}{|A_\ast|} \sum_{w_k \in A_\ast} \widetilde{u}_{A,k},
\end{equation}
where $\widetilde{u}_{A,k}$ is the utility of selecting subset $A$ when the true class is $k$, and we normalize this average utility to ensure that $\widetilde{u}_{A_\ast,A_\ast} = 1$:
\begin{equation}
\widetilde{u}_{A,A_\ast}=\frac{\overline{u}_{A,A_\ast}}{\overline{u}_{A_\ast,A_\ast}}.
\end{equation}
\end{subequations}

\begin{Ex}\label{ex:softlable}
 Table \ref{tab:example_utility_softlabel} shows an example of the utility matrix considering soft labels, which is extended from Example \ref{ex:nosoftlable}. The last four columns correspond to the utility matrix for soft labels. An act achieves utility 1 only if $A=A_\ast$, 0 if $A \cap A_\ast=\emptyset$, and a value between 0 and 1 if $A\neq A_\ast$ and $A \cap A_\ast \neq \emptyset$. 
 
 \begin{table}[]
  \centering
  \caption{Utility matrix considering soft labels with $\gamma=0.8$.}\label{tab:example_utility_softlabel}
  \resizebox{\linewidth}{!}{
   \begin{tabular}{ccccccccc}
    \hline
    \multicolumn{2}{c}{\multirow{2}{*}{}}        & \multicolumn{7}{c}{Label}                                                                       \\ \cline{3-9} 
    \multicolumn{2}{c}{}                         & $\omega_1$ & $\omega_2$ & $\omega_3$ & $\{\omega_1,\omega_2\}$ & $\{\omega_1,\omega_3\}$ & $\{\omega_2,\omega_3\}$ & $\Omega$ \\ \hline
    \multirow{7}{*}{Act} & $f_{\{\omega_1\}}$    & 1          & 0          & 0          & 0.625        & 0.625       & 0             & 0.489    \\
    & $f_{\{\omega_2\}}$    & 0          & 1          & 0          & 0.625         & 0             & 0.625         & 0.489    \\
    & $f_{\{\omega_3\}}$    & 0          & 0          & 1          & 0             & 0.625         & 0.625         & 0.489    \\
    & $f_{\{\omega_1,\omega_2\}}$ & 0.8        & 0.8        & 0          & 1             & 0.5           & 0.5           & 0.782    \\
    & $f_{\{\omega_1,\omega_3\}}$ & 0.8        & 0          & 0.8        & 0.5           & 1             & 0.5           & 0.782    \\
    & $f_{\{\omega_2,\omega_3\}}$ & 0          & 0.8        & 0.8        & 0.5           & 0.5           & 1             & 0.782    \\
    & $f_{\Omega}$      & 0.682      & 0.682      & 0.682      & 0.853         & 0.853         & 0.853         & 1        \\ \hline
  \end{tabular}}
 \end{table} 
 
\end{Ex}

\paragraph{Utility of intersection over union.} The segmentation performance was also evaluated by \emph{the utility of intersection over union} (UIoU) defined as
\begin{equation}\label{con:uiou}
UIoU=\frac{1}{2^{|\Omega|}-1}\sum_{B \subseteq \Omega}\frac{\sum_{i \in \boldsymbol G^B \cap \boldsymbol P^B}{\widetilde u}_{A(i),B}}{|\boldsymbol G^B \cup \boldsymbol P^B|},
\end{equation}
where $\boldsymbol P^B=\{i: A(i) \cap B\neq \emptyset\}$ is the predicted area containing pixels assigned to a set of classes that intersect $B$, and $\boldsymbol G^B=\{i: A_*(i)=B)\}$ is the ground truth area composed of pixels with label $B$. Thus, in the special case of precise segmentation with only precise labels, UIoU boils down to \emph{intersection over union}, a widely used metric for semantic segmentation \cite{long2015fully,noh2015learning,krahenbuhl2011efficient}.

\paragraph{Expected calibration error.} In decision systems, a neural network should not only be accurate, but it should also indicate when it is likely to be incorrect. Thus, the confidence of an E-FCN should be \emph{calibrated}. To characterize this property, we extend the   \emph{expected calibration error} (ECE) defined in \cite{guo2017calibration} as follows. We define the \emph{prediction confidence} of pixel $i$  as
\begin{equation}\label{con:pconfidence}
co(i)=BetP_{i}(A_\ast(i))=\sum_{\omega_j \in A_\ast(i)}BetP_{i}(\{\omega_j\}) ,
\end{equation}
where $BetP_{i}$ is the predicted pignistic probability measure for pixel $i$. Let $I_q$ be the set  of pixels whose prediction confidence lies in the interval $(\frac{q-1}{Q}, \frac{q}{Q}]$, $q=1,\dots,Q$. The average utility and confidence of $I_q$ are defined, respectively, as
\begin{subequations}\label{con:ucbin}
 \begin{equation}\label{con:ubin}
 au(I_q)=\frac{1}{|I_q|}\sum_{i \in I_q} {\widetilde u}_{A(i),A_\ast(i)},
 \end{equation}
 and
 \begin{equation}\label{con:cbin}
 co(I_q)=\frac{1}{|I_q|} \sum_{i \in I_q} co(i).
 \end{equation}
\end{subequations}
We consider that the classifier is well calibrated if $co(I_q)\approx au(I_q)$ for all $q$, and we define  the ECE  as
 \begin{equation}\label{con:sum_ece}
 ECE=\frac{\sum_{q=1}^Q |I_q| \times |co(I_q)-au(I_q)|}{\sum_{q'=1}^Q |I_q'|}
 \end{equation}
When only considering precise acts and labels,  ECE defined by \eqref{con:sum_ece} boils down to the original definition in \cite{guo2017calibration}.

\subsection{Precise segmentation results}
\label{sec:precise_segmentation}

In precise segmentation, each pixel of an image is assigned to exactly one class,  the set of acts being defined as $\mathcal{F}=\{f_{\omega_1},\dots,f_{\omega_M}\}$. Three databases without soft labels mentioned in Section \ref{sec:databases} were used to train and test the E-FCNs and probabilistic FCNs (P-FCNs). The metrics defined in Section \ref{sec:metrics} with the utility matrix $\bU$ equal to the identity matrix were used for  performance assessment. 

\begin{figure}
	\centering
	\subfloat[\label{fig:illustration_fcn_s}]{\includegraphics[width=\textwidth]{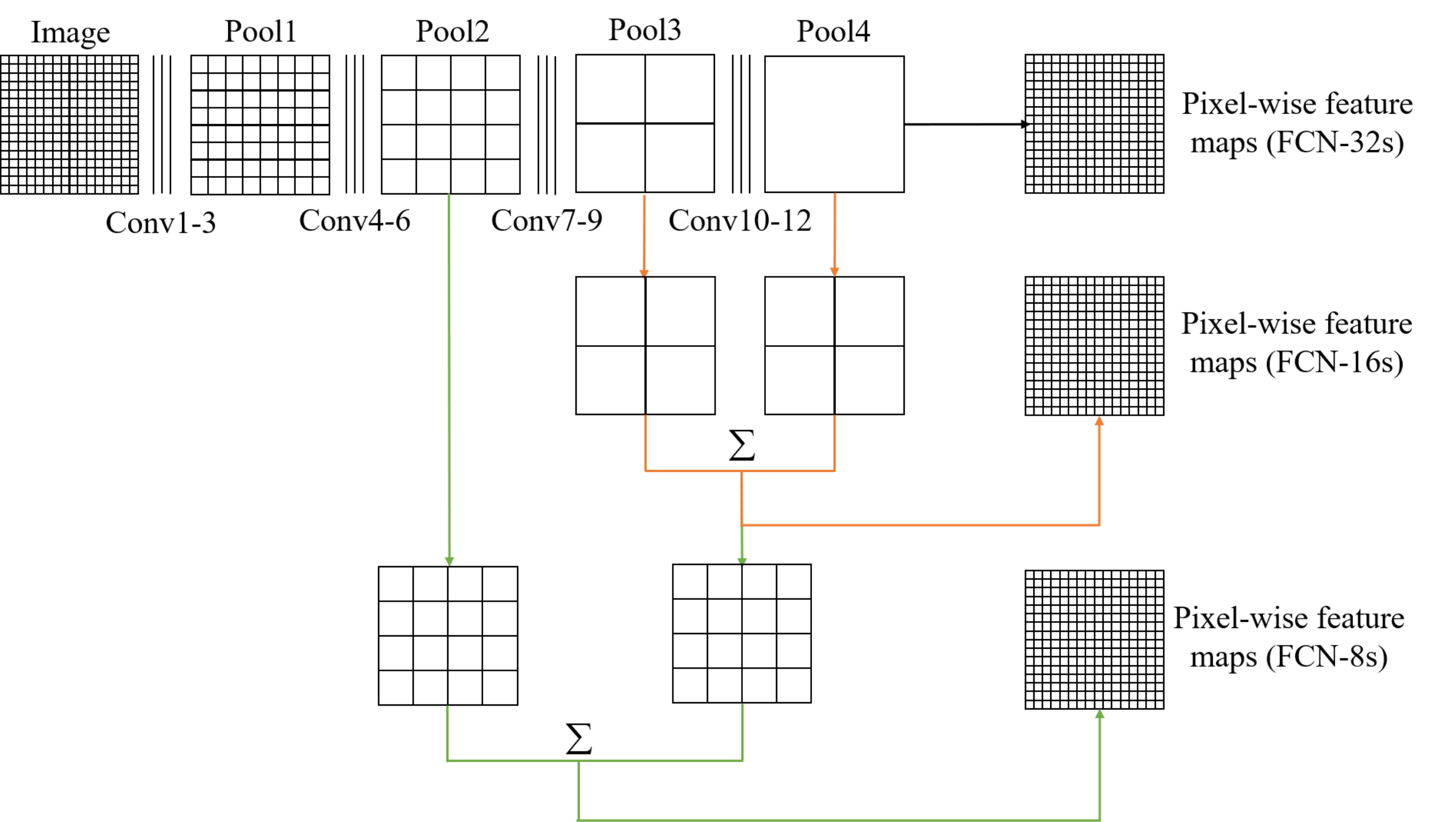}}\\
	\subfloat[\label{fig:illustration_fcn_seg}]{\includegraphics[width=0.7\textwidth]{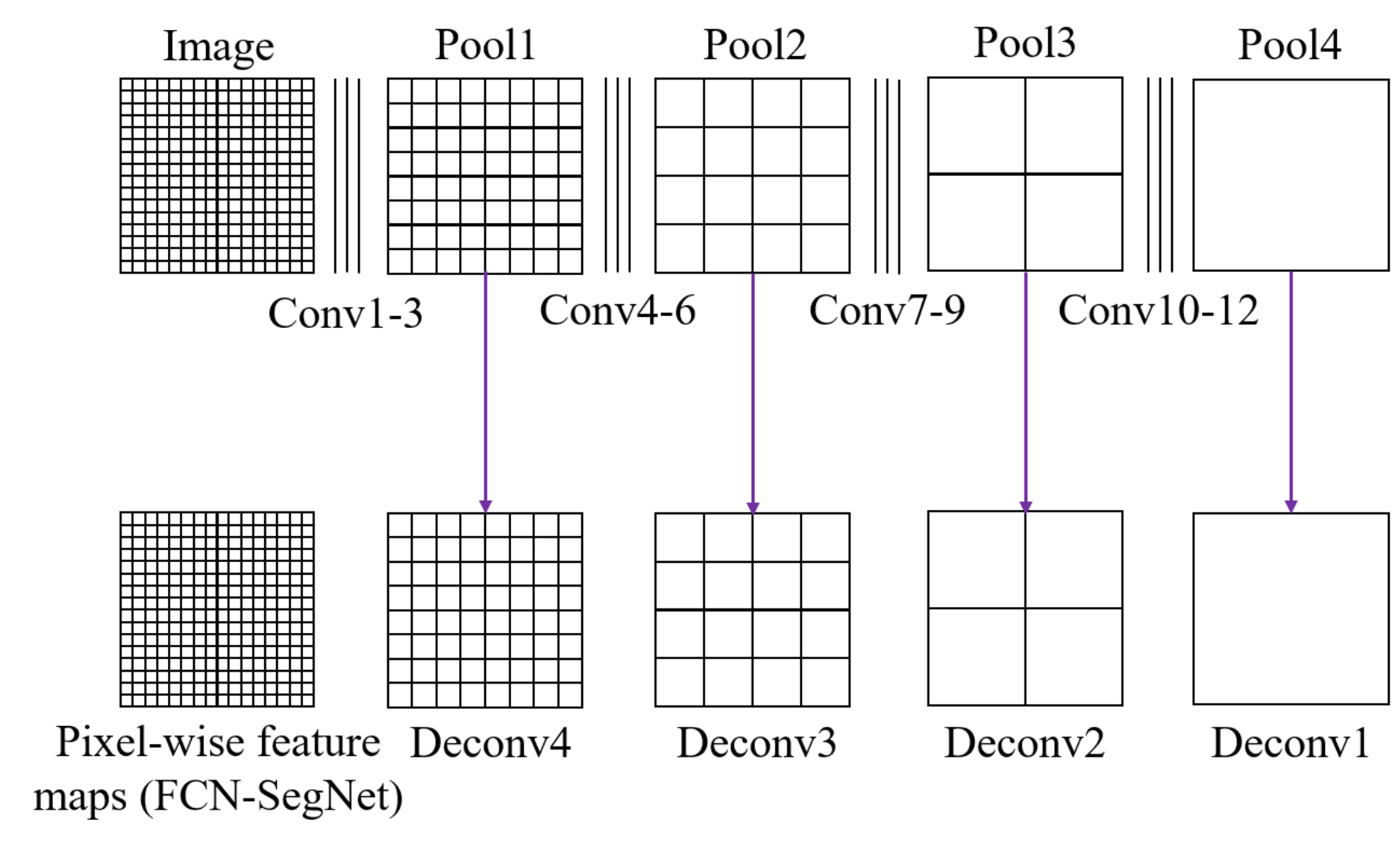}}\\
	\caption{Illustration of the encoder-decoder architectures used in this paper. Pooling layers are represented as grids that show relatively sparse information. Intermediate convolution layers are omitted. (a) The FCN-32s, FCN-16s, FCN-8s architectures are used to combine sparse and high-layer information with dense and low-layer information for upsampling.  Black arrow: the deconvolutional layer in FCN-32s directly upsamples the outputs of Pool 4 to pixel-wise feature maps; orange arrows: the deconvolutional layer in FCN-16s combines outputs from Pool 3 and 4, lets the net predict finer details, while retaining high-level semantic information; green arrows: the deconvolutional layer in FCN-8s acquire additional feature maps from Pool 2 to provide further precision; (b) The FCN-SegNet architecture uses four deconvolutional layers to upsample the sparse feature maps from the end of the encoder part, as well as the feature maps from the corresponding pooling layers based on pooling indices (purple arrows).}\label{fig:illustration_fcn}
\end{figure}

In the experiment with each database, three widely used encoder-decoder architectures were combined with the DS and utility layers, as shown in Table \ref{tab:precise_results}. All encoder-decoder architectures in Table \ref{tab:precise_results} have the same encoder part, which consists of four stages and two convolutional layers with $3 \times 3$ kernels. Each stage is made up of three convolutional layers with $3 \times 3$ kernels and a max-pooling layer with a $2 \times 2$ non-overlapping window. Figure \ref{fig:illustration_fcn_s} illustrates the differences between the FCN-32s, FCN-16s, and FCN-8s architectures in their decoder parts with a deconvolutional layer. The FCN-SegNet architecture uses four deconvolutional layers to upsample the sparse feature maps from the end of the encoder part, as well as the feature maps from the corresponding pooling layers based on pooling indices \cite{badrinarayanan2017segnet}, as shown in Figure \ref{fig:illustration_fcn_seg}. The FCN-DilatedVGG architecture is the same as  FCN-SegNet  except that it adds a fully connected conditional random field at the end of the last deconvolutional layer \cite{chen2017deeplab}. The numbers $P$ of feature maps for the Pascal, MIT and SIFT databases were, respectively, 31, 128 and 64. The numbers $n$ of prototypes in the DS layer for these three databases were set, respectively, to 75, 300 and 95.

\begin{table}
 \caption{Performance evaluation of precise segmentation: (a) Pascal VOC 2011, (b) MIT-scene Parsing, and (c) SIFT Flow. P-FCN and E-FCN are, respectively, probabilistic and evidential FCNs. The rests of the notations, such as ``-32s'' and ``-16s'',  stand for different encoder-decoder architectures.  The results are in form of ``mean value $\pm$ standard deviation''. The best results for each encoder-decoder architecture are highlighted in bold.}\label{tab:precise_results}
 \centering
 \subfloat[\label{tab:voc_precise}]{
  \begin{tabular}{lcc}
   \hline
   & PU             & UIoU           \\ \hline
   P-FCN-32s \cite{long2015fully} & 0.8912 $\pm$ 0.0019         & 0.5941 $\pm$ 0.0033    \\
   P-FCN-16s \cite{long2015fully} & 0.9001 $\pm$ 0.0015         & 0.6243 $\pm$ 0.0025         \\
   P-FCN-8s  \cite{long2015fully} & 0.9033 $\pm$ 0.0017         & 0.6269 $\pm$ 0.0021         \\ \hline
   E-FCN-32s  & \textbf{0.8973} $\pm$ 0.0021 & \textbf{0.6128} $\pm$ 0.0024 \\
   E-FCN-16s & \textbf{0.9045} $\pm$ 0.0014 & \textbf{0.6304} $\pm$ 0.0019 \\
   E-FCN-8s   & \textbf{0.9074} $\pm$ 0.0015 & \textbf{0.6337} $\pm$ 0.0020 \\ \hline
 \end{tabular}}
 \qquad
 \subfloat[\label{tab:mit_precise}]{
  \begin{tabular}{lcc}
   \hline
   & PU             & UIoU           \\ \hline
   P-FCN-16s \cite{long2015fully}       & 0.7009 $\pm$ 0.0030        & 0.289  $\pm$ 0.0051        \\
   P-FCN-8s  \cite{long2015fully}       & 0.7128 $\pm$ 0.0024        & 0.294  $\pm$ 0.0048        \\
   P-FCN-SegNet \cite{badrinarayanan2017segnet}   & 0.7153 $\pm$ 0.0023        & 0.305   $\pm$ 0.0042       \\ \hline
   E-FCN-16s         & \textbf{0.7090} $\pm$ 0.0026 & \textbf{0.292} $\pm$ 0.0048\\
   E-FCN-8s         & \textbf{0.7148} $\pm$ 0.0025 & \textbf{0.296} $\pm$ 0.0046\\
   E-FCN-SegNet   & \textbf{0.7167} $\pm$ 0.0026 & \textbf{0.330}  $\pm$ 0.0043\\ \hline
 \end{tabular}}
 \qquad
 \subfloat[\label{tab:sift_precise}]{
  \begin{tabular}{lcc}
   \hline
   & PU             & UIoU           \\ \hline
   P-FCN-16s  \cite{long2015fully}   & 0.8489 $\pm$ 0.0034          & 0.3922  $\pm$ 0.0047        \\
   P-FCN-8s   \cite{long2015fully}   & 0.8525 $\pm$ 0.0032         & 0.3948   $\pm$ 0.0042       \\
   P-FCN-DilatedVGG \cite{chen2017deeplab} & 0.8643 $\pm$ 0.0036          & 0.4168    $\pm$ 0.0043      \\ \hline
   E-FCN-16s      & \textbf{0.8521} $\pm$ 0.0030 & \textbf{0.3937}$\pm$ 0.0042 \\
   E-FCN-8s      & \textbf{0.8528} $\pm$ 0.0031 & \textbf{0.3961} $\pm$ 0.0040\\
   E-FCN-DilatedVGG  & \textbf{0.8649} $\pm$ 0.0035 & \textbf{0.4182} $\pm$ 0.0038\\ \hline
 \end{tabular}}
\end{table}

The DS and utility layers slightly improve the accuracy of precise assignments performed by FCN models, even though the performance of FCN models on precise segmentation mainly depends on the encoder-decoder architecture. Table \ref{tab:voc_precise} presents the results of PU and UIoU for the Pascal VOC database. E-FCNs achieved higher PU and UIoU than P-FCNs with the same encoder-decoder architecture, which shows the E-FCNs outperform the P-FCNs for precise segmentation. Similar improvements can also be found in the MIT-scene Parsing and SIFT Flow databases as shown, respectively, in Tables \ref{tab:mit_precise} and  \ref{tab:sift_precise}. 
 
\begin{figure}
 \centering
 \includegraphics[width=\textwidth]{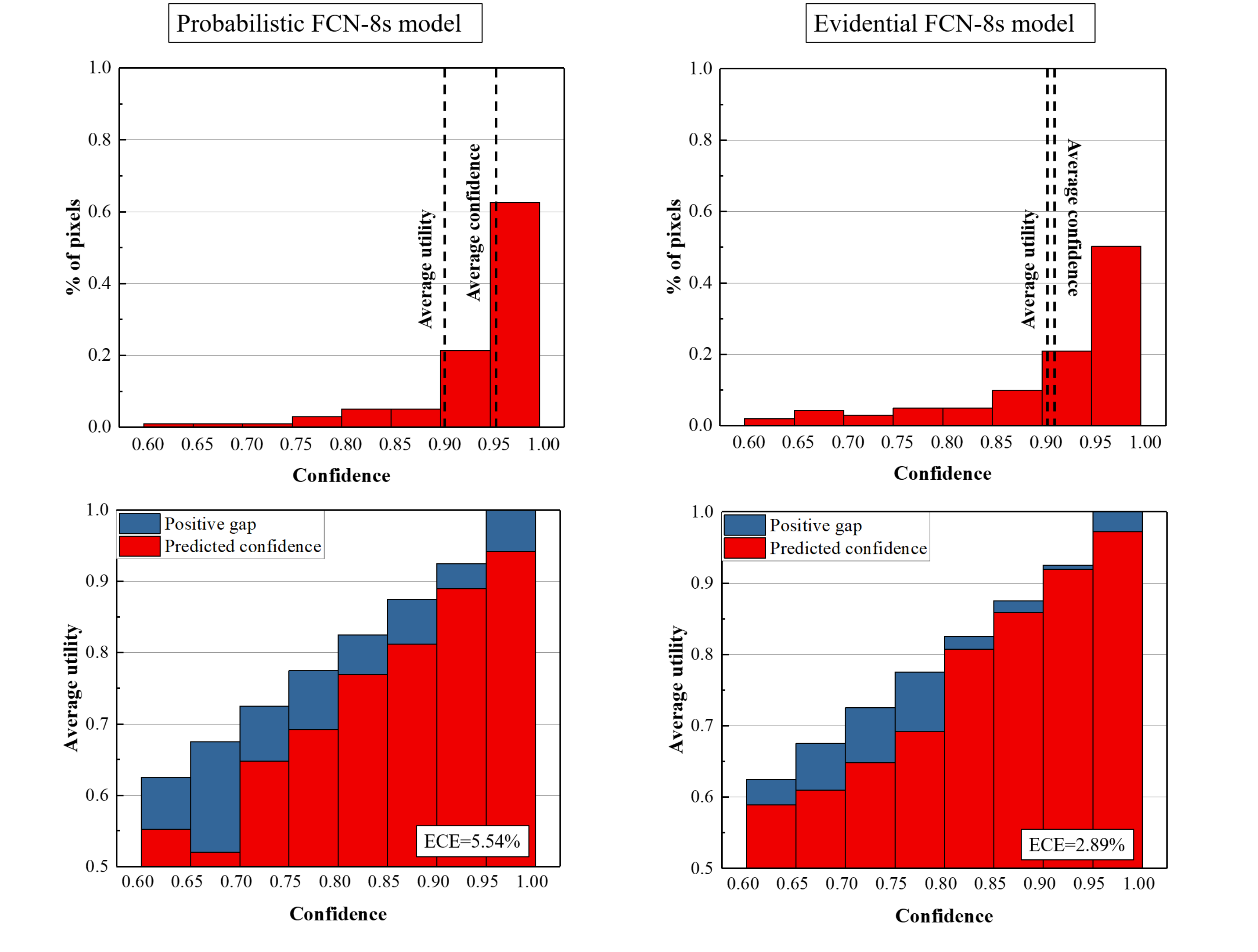}\\
 \caption{Pixel confidence distributions (top) and pixel utility histograms (bottom) for P-FCN-8s (left) and E-FCN-8s (right) on the Pascal VOC database.}\label{fig:ece_precise_voc}
\end{figure}

\begin{figure}
 \centering
 \includegraphics[width=0.9\textwidth]{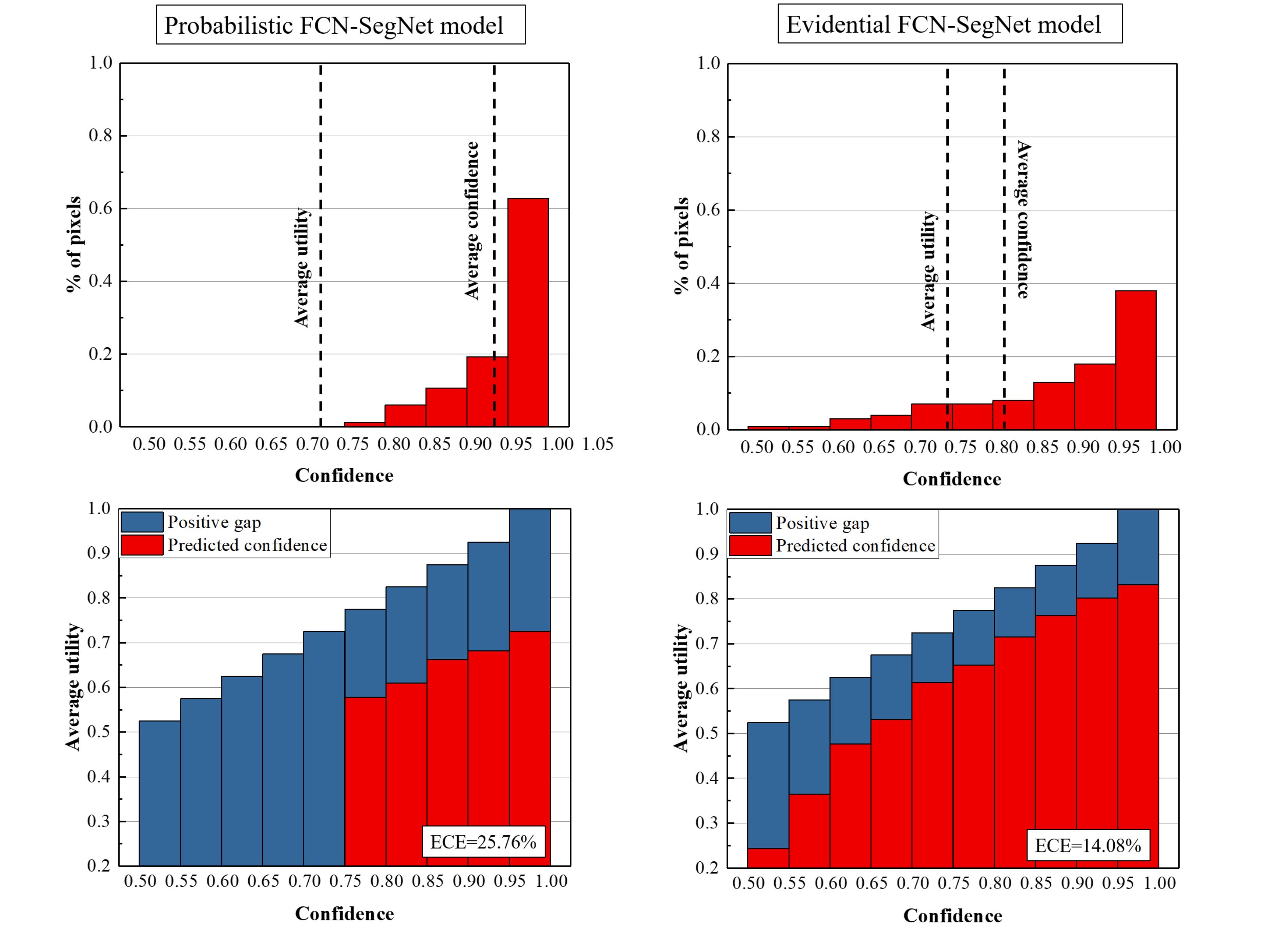}\\
 \caption{Pixel confidence distributions (top) and pixel utility histograms (bottom) for P-FCN-SegNet (left) and E-FCN-SegNet (right) on the MIT-scene Parsing database.}\label{fig:ece_precise_mit}
\end{figure}

\begin{figure}
 \centering
 \includegraphics[width=0.9\textwidth]{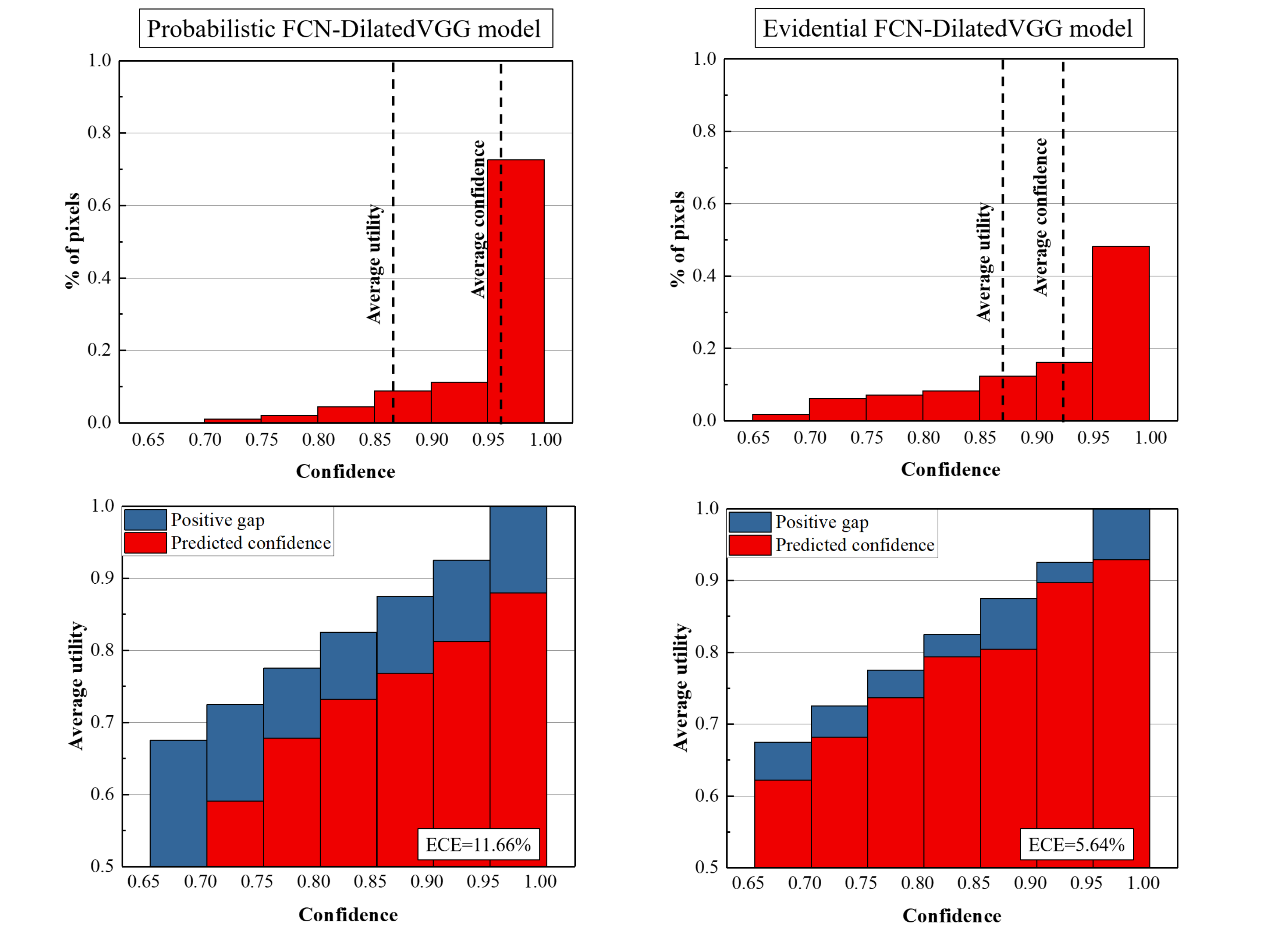}\\
 \caption{Pixel confidence distributions (top) and pixel utility histograms (bottom) for P-FCN-DilatedVGG (left) and E-FCN-DilatedVGG (right) on the SIFT Flow database.}\label{fig:ece_precise_sift}
\end{figure}

The use of DS and utility layers also makes the FCN models better calibrated. Figure \ref{fig:ece_precise_voc} presents a visual calibration representation of the FCN-8s models in the Pascal VOC database. The top row shows the pixel distribution of prediction confidence \eqref{con:cbin} as histograms. The average confidence of the E-FCN-8s model closely matches its average pixel utility, while the average confidence of the P-FCN-8s model is substantially higher than its average pixel utility. This is further illustrated in the bottom row of pixel utility diagrams, which show pixel utility as a function of confidence. The E-FCN-8s model is well calibrated since its confidence in each bin approximates the expected average utility, whereas the predicted utility of the P-FCN-8s model does not match its confidence. As a consequence, the E-FCN-8s model achieves a smaller ECE than the probabilistic one. The effect of the DS and utility layers on the calibration can also be found in the FCN-SegNet and FCN-DialtedVGG models on the MIT-scene Parsing and SIFT Flow databases as shown, respectively, in Figures \ref{fig:ece_precise_mit} and  \ref{fig:ece_precise_sift}.

\subsection{Imprecise segmentation results}
\label{sec:imprecise_segmentation}

In  imprecise segmentation, each pixel of an image is assigned to a non-empty subset $A$ of $\Omega$; the set of acts is $\mathcal{F}=\{f_A,A \in 2^\Omega \backslash {\emptyset}\}$, or a subset thereof. Here we only considered acts $f_A$ such that $A$ is a singleton, $\Omega$ or one of the soft labels in the training set. For performance evaluation, we used  the metrics and the  three databases described  in Section \ref{sec:databases_metrics}. For each database, the segmentation masks with and without soft labels were used to train different FCN models. The same  encoder-decoder architectures used for precise segmentation  in Section \ref{sec:precise_segmentation} were combined with the DS and utility layers.

\begin{figure}
 \centering
 \subfloat[\label{fig:pu_nosoft_voc}]{\includegraphics[width=0.45\textwidth]{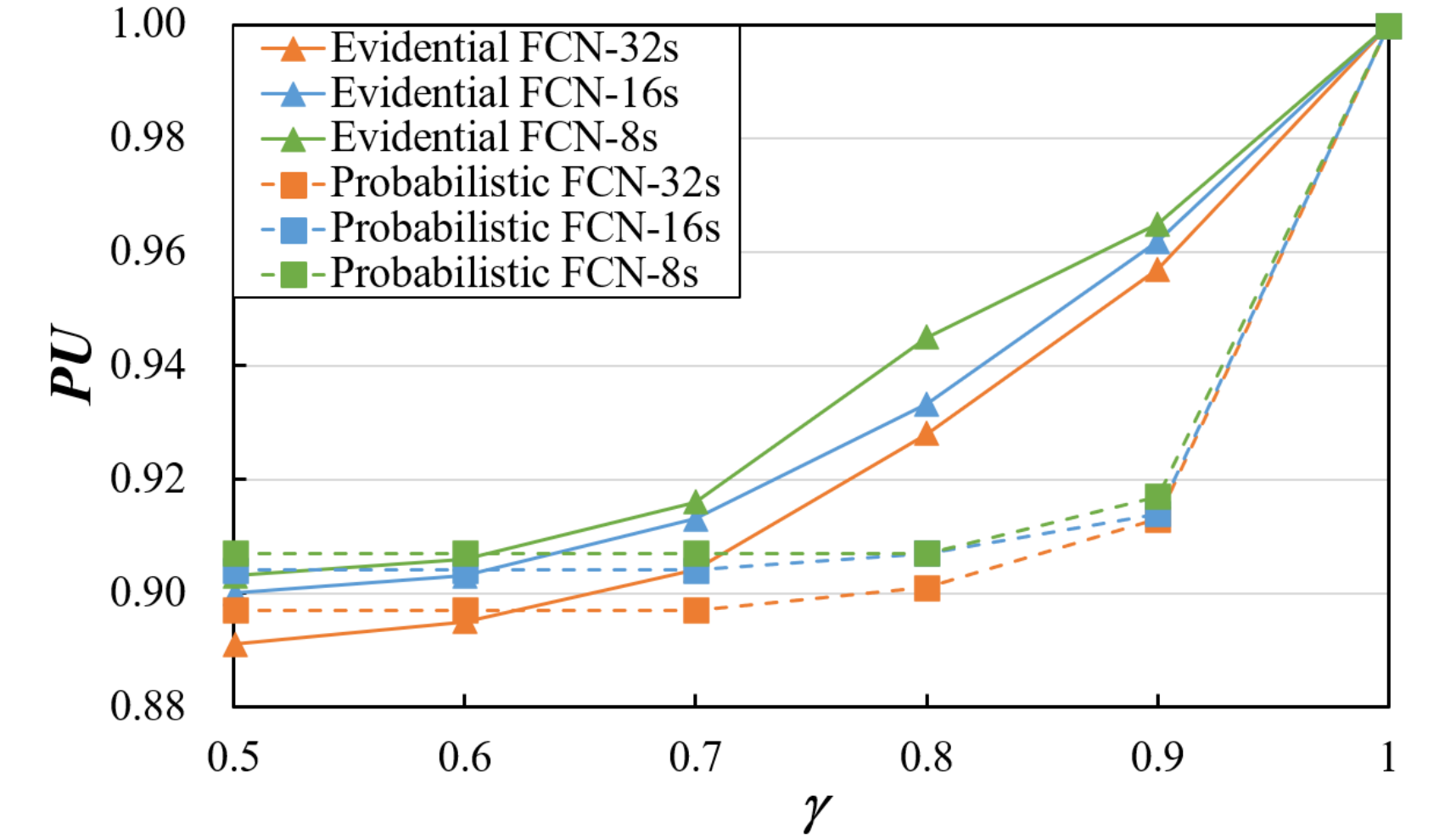}}\hspace{0.01\textwidth}
 \subfloat[\label{fig:pu_soft_voc}]{\includegraphics[width=0.45\textwidth]{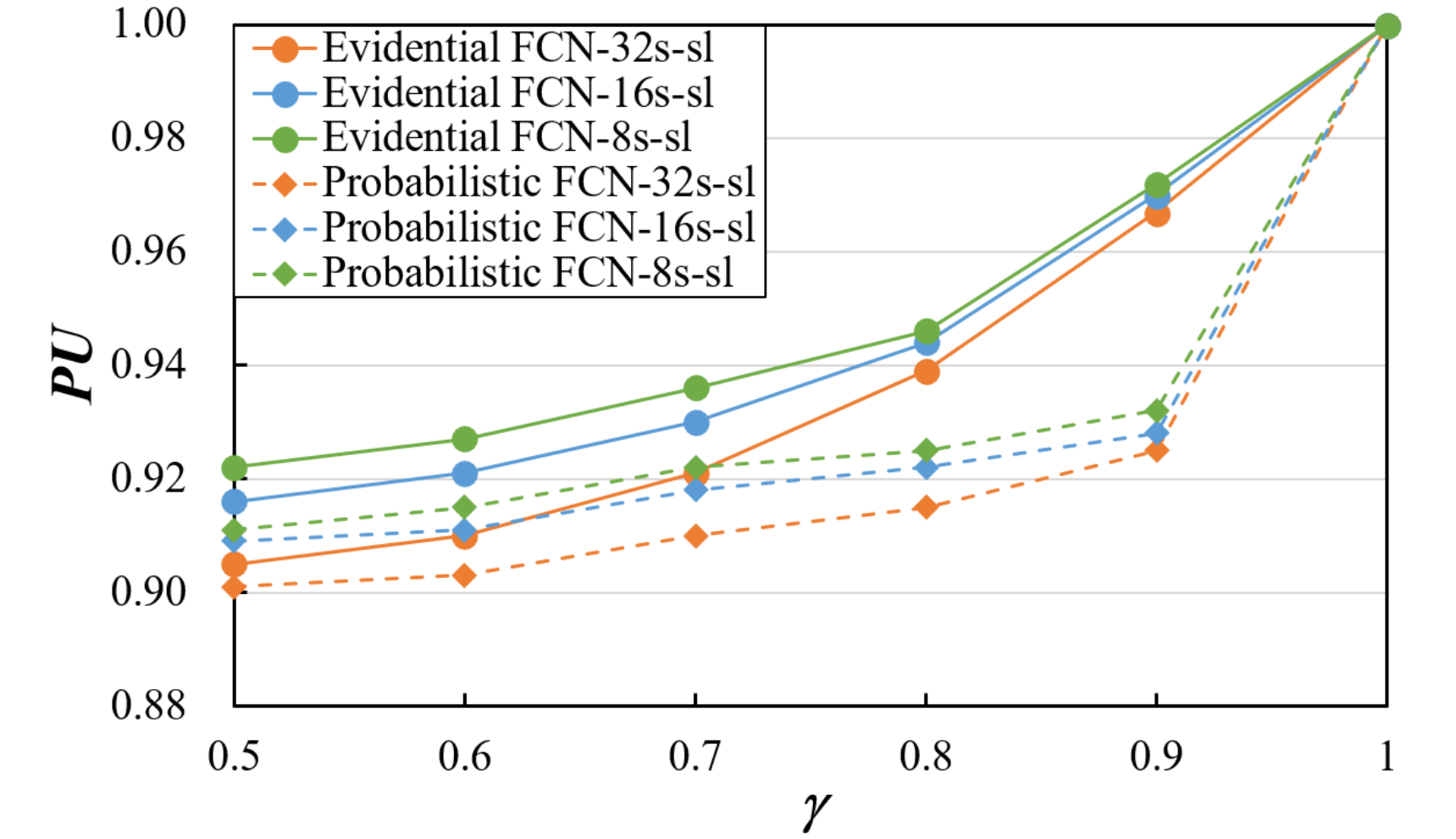}}\\
 \subfloat[\label{fig:uiou_nosoft_voc}]{\includegraphics[width=0.45\textwidth]{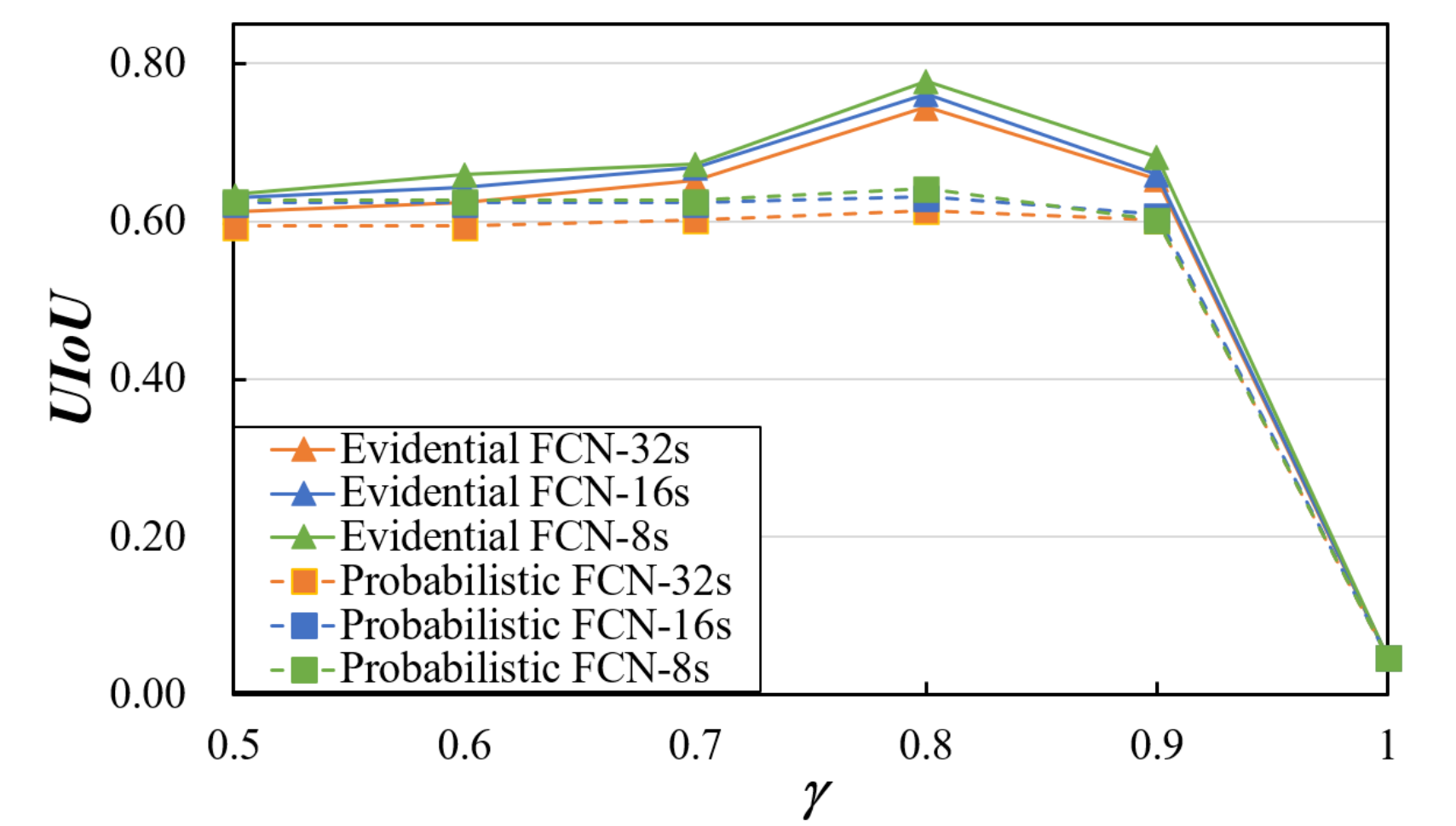}}\hspace{0.01\textwidth}
 \subfloat[\label{fig:uiou_soft_voc}]{\includegraphics[width=0.45\textwidth]{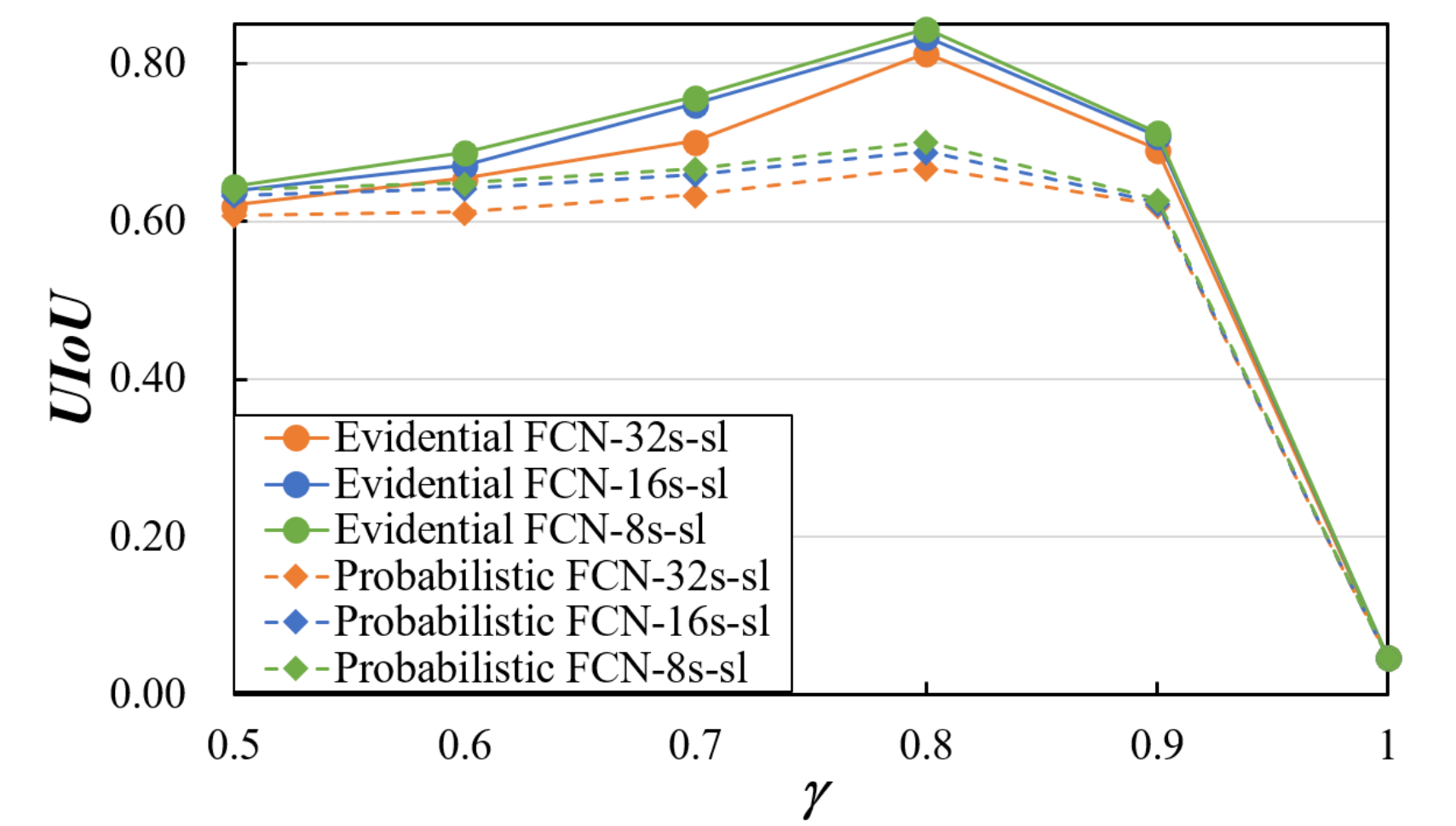}}\hspace{0.01\textwidth}
 \caption{Testing PU and UIoU vs. $\gamma$ on the Pascal VOC database. The first and second columns are the models trained with/without soft labels, respectively.}\label{fig:pu_uiou_imprecise_segmentation_voc}
\end{figure}

\begin{figure}
 \centering
 \includegraphics[width=0.85\textwidth]{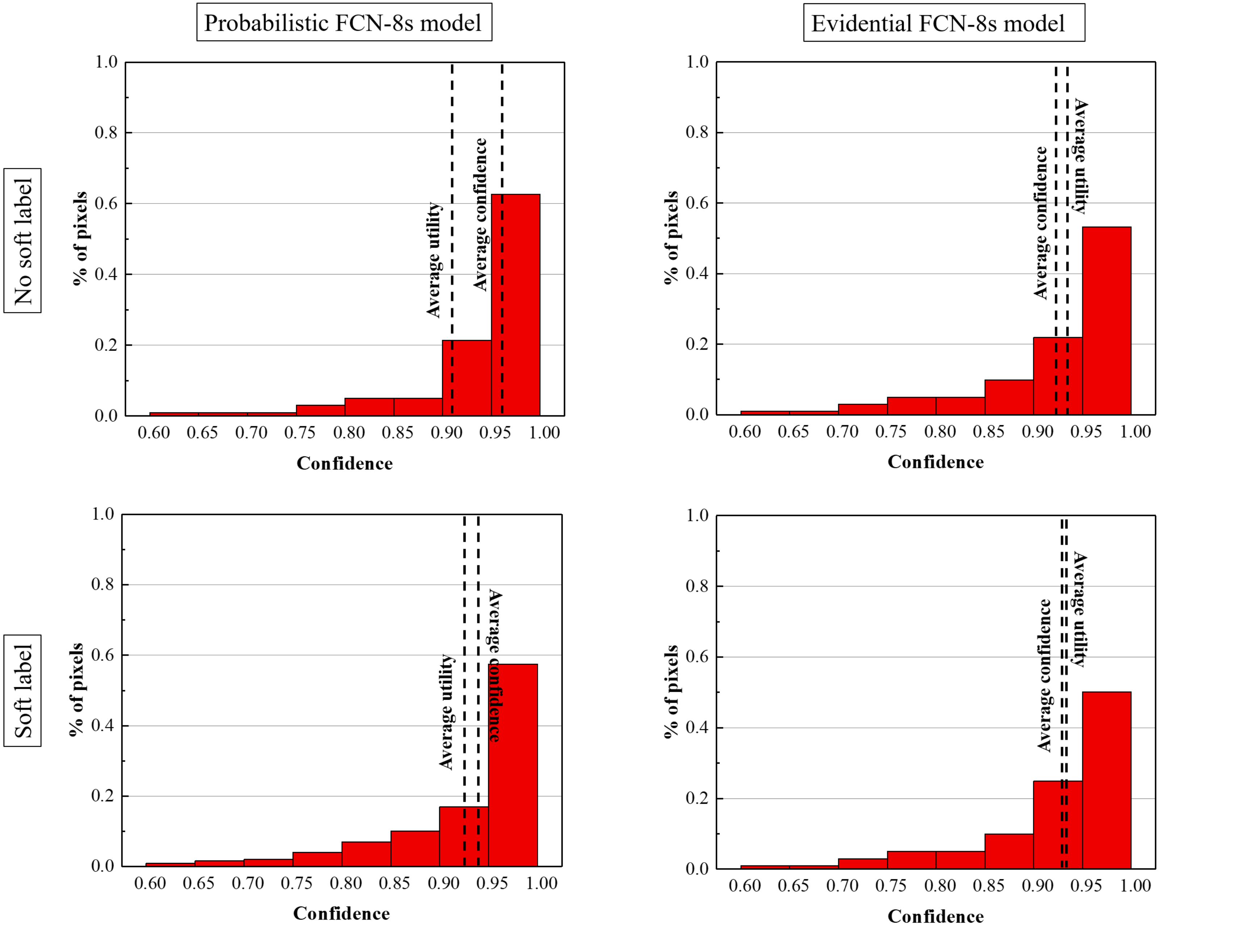}
 \caption{Pixel confidence distributions for the P-FCN-8s (left) and E-FCN-8s (right) models on the Pascal VOC 2011 database without (top)/with (bottom) soft labels.}\label{fig:pixeldistribution_voc}
\end{figure}

\begin{figure}
 \centering
 \subfloat[\label{fig:pu_nosoft_mit}]{\includegraphics[width=0.49\textwidth]{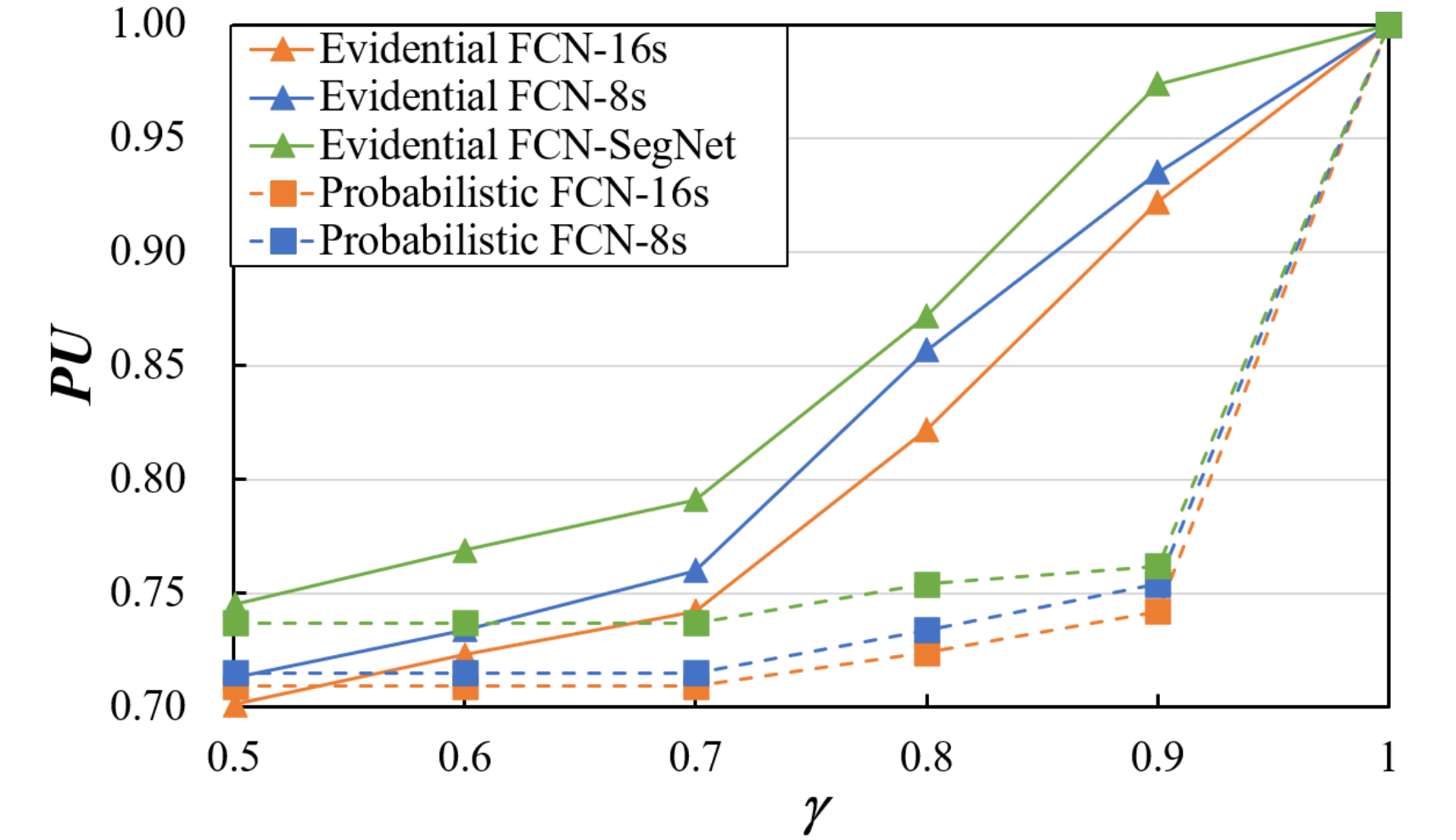}}\hspace{0.01\textwidth}
 \subfloat[\label{fig:pu_soft_mit}]{\includegraphics[width=0.49\textwidth]{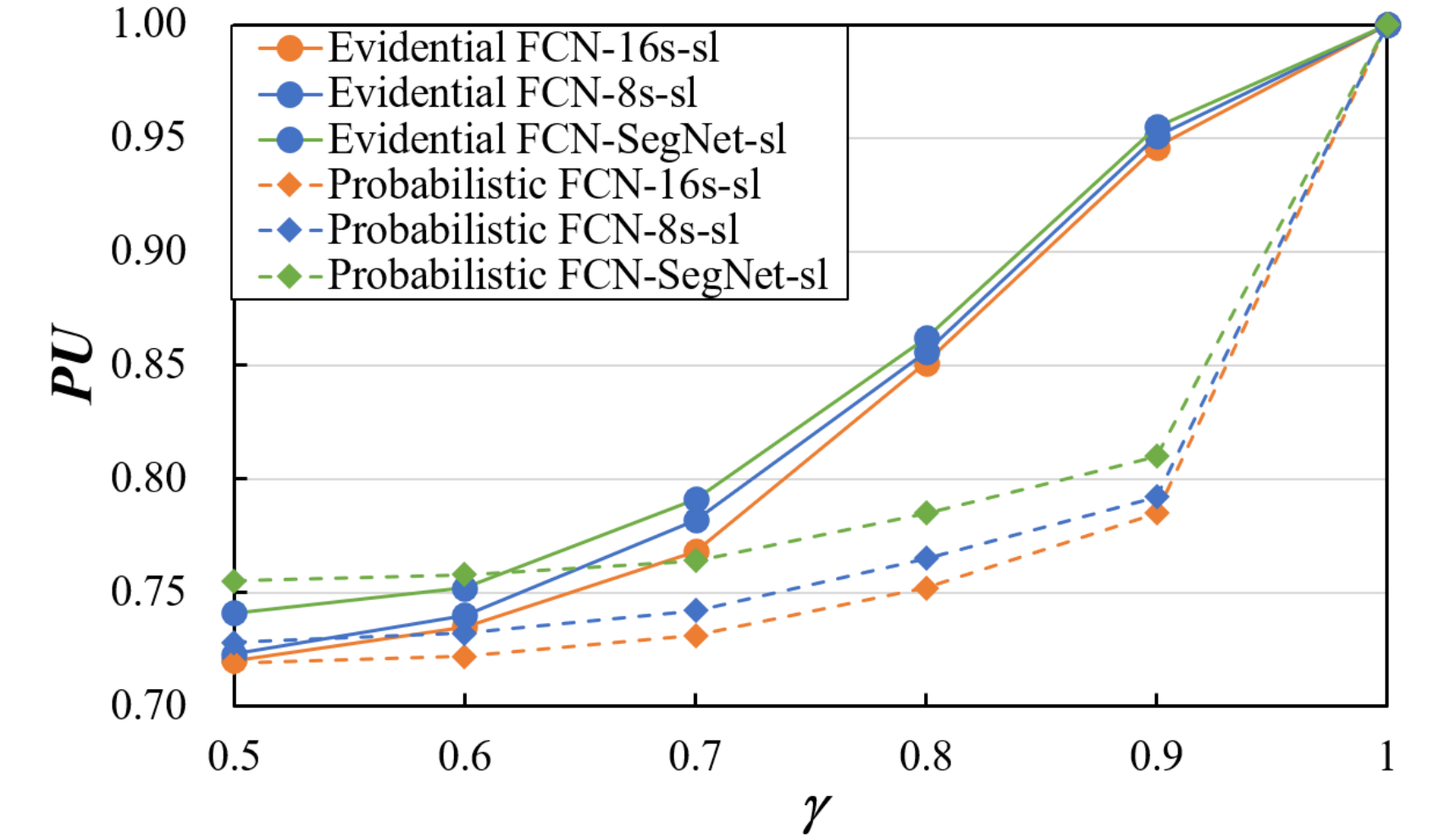}}\\
 \subfloat[\label{fig:uiou_nosoft_mit}]{\includegraphics[width=0.49\textwidth]{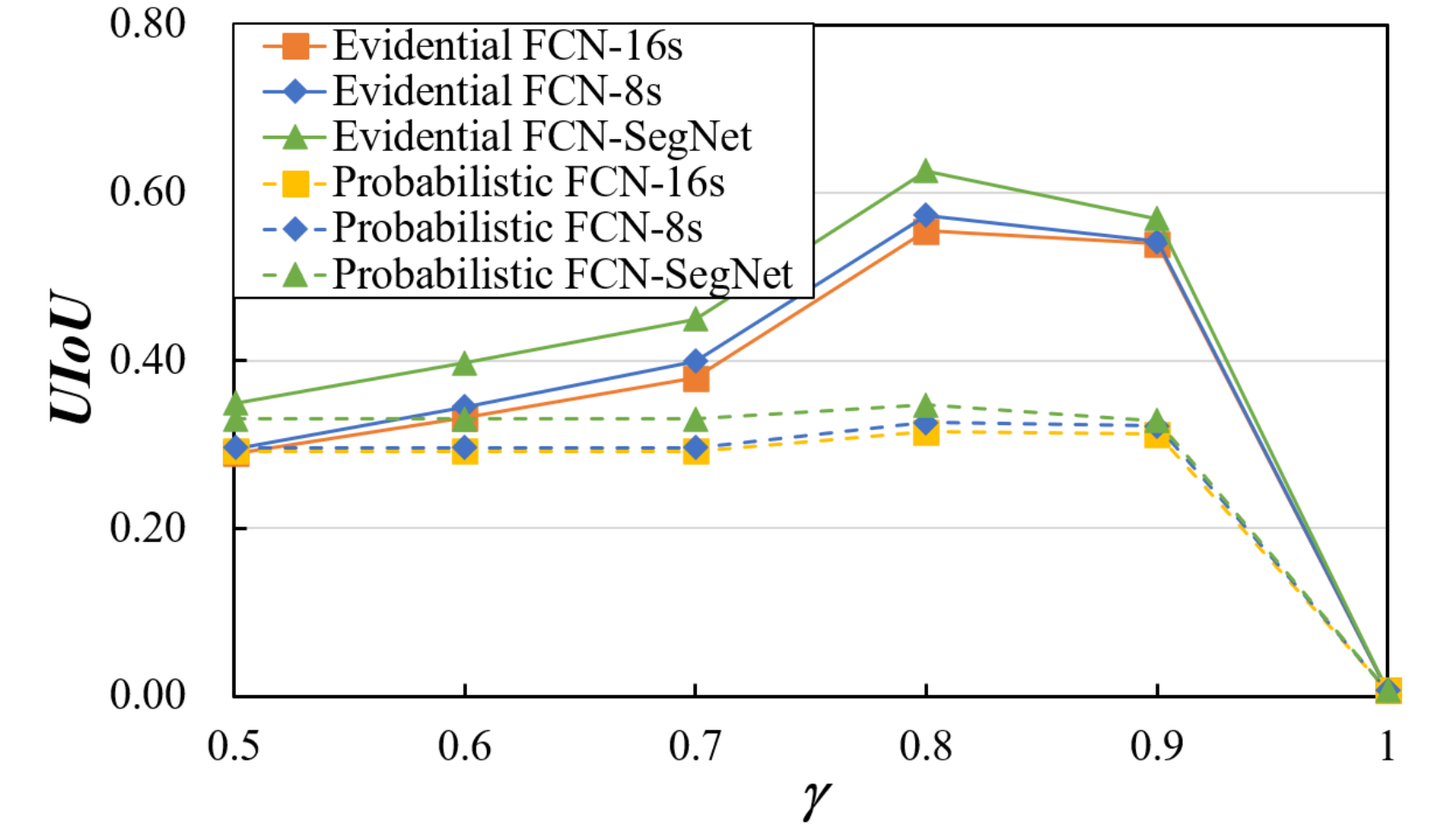}}\hspace{0.01\textwidth}
 \subfloat[\label{fig:uiou_soft_mit}]{\includegraphics[width=0.49\textwidth]{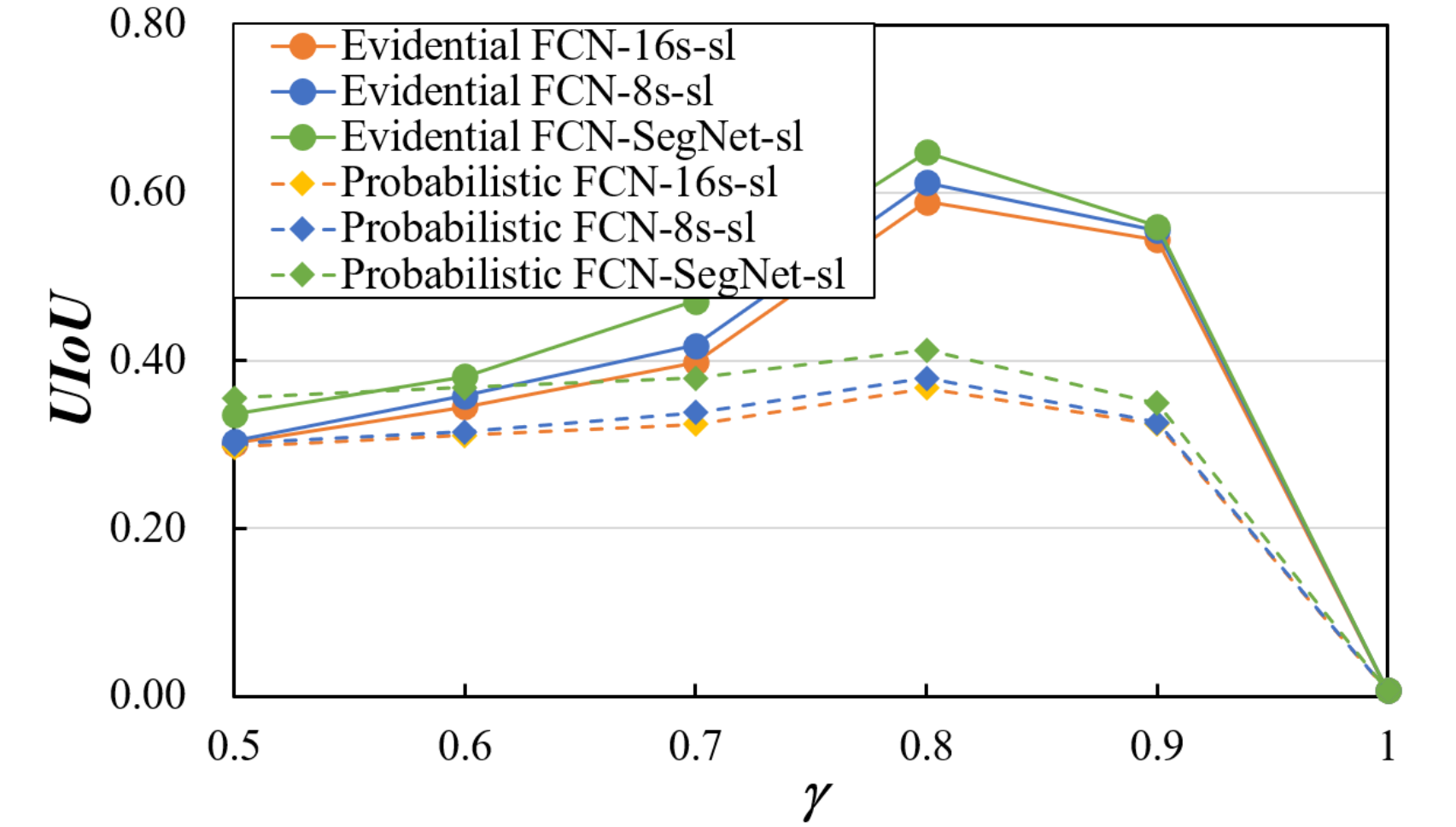}}\hspace{0.01\textwidth}
 \caption{Testing PU and UIoU vs. $\gamma$ on the MIT-scene Parsing database. The first and second columns are the models trained with/without soft labels, respectively.}\label{fig:pu_uiou_imprecise_segmentation_mit}
\end{figure}

\begin{figure}
 \centering
 \includegraphics[width=0.85\textwidth]{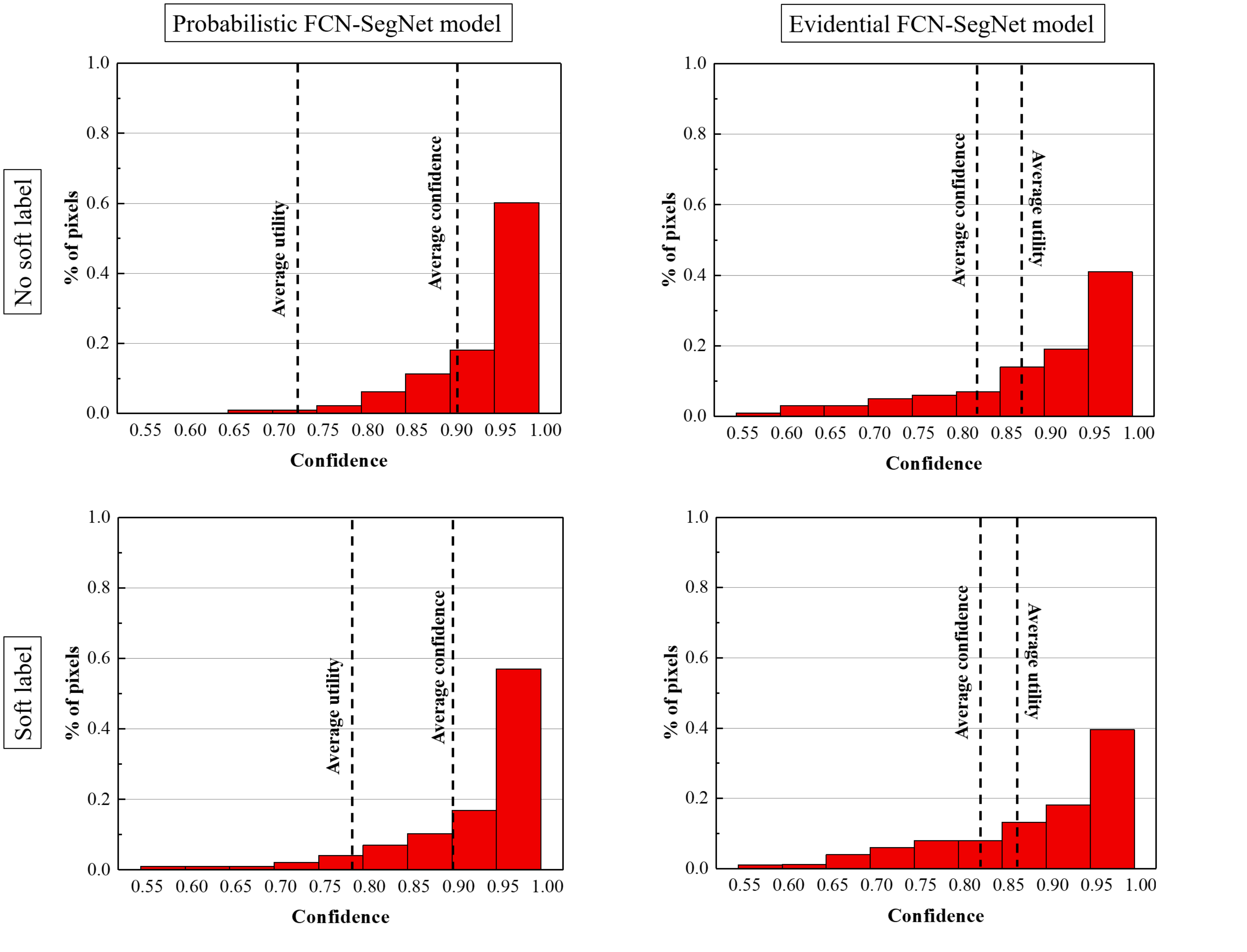}
 \caption{Pixel rate histograms for the P-FCN-SegNet (left) and E-FCN-SegNet (right) models on the MIT-scene Parsing database without (top)/with (bottom) soft labels.}\label{fig:pixeldistribution_mit}
\end{figure}

\begin{figure}
 \centering
 \subfloat[\label{fig:pu_nosoft_sift}]{\includegraphics[width=0.49\textwidth]{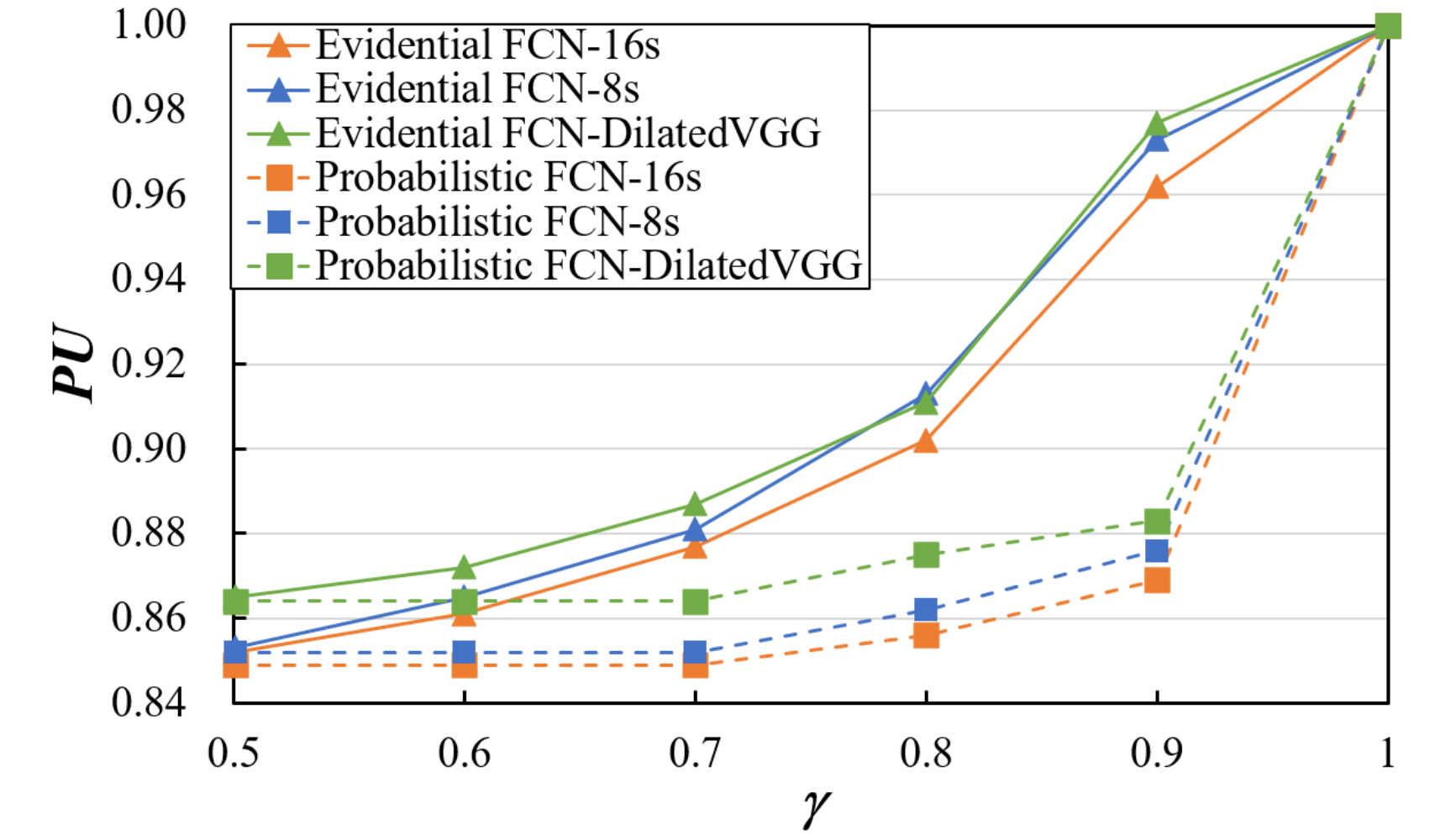}}\hspace{0.01\textwidth}
 \subfloat[\label{fig:pu_soft_sift}]{\includegraphics[width=0.49\textwidth]{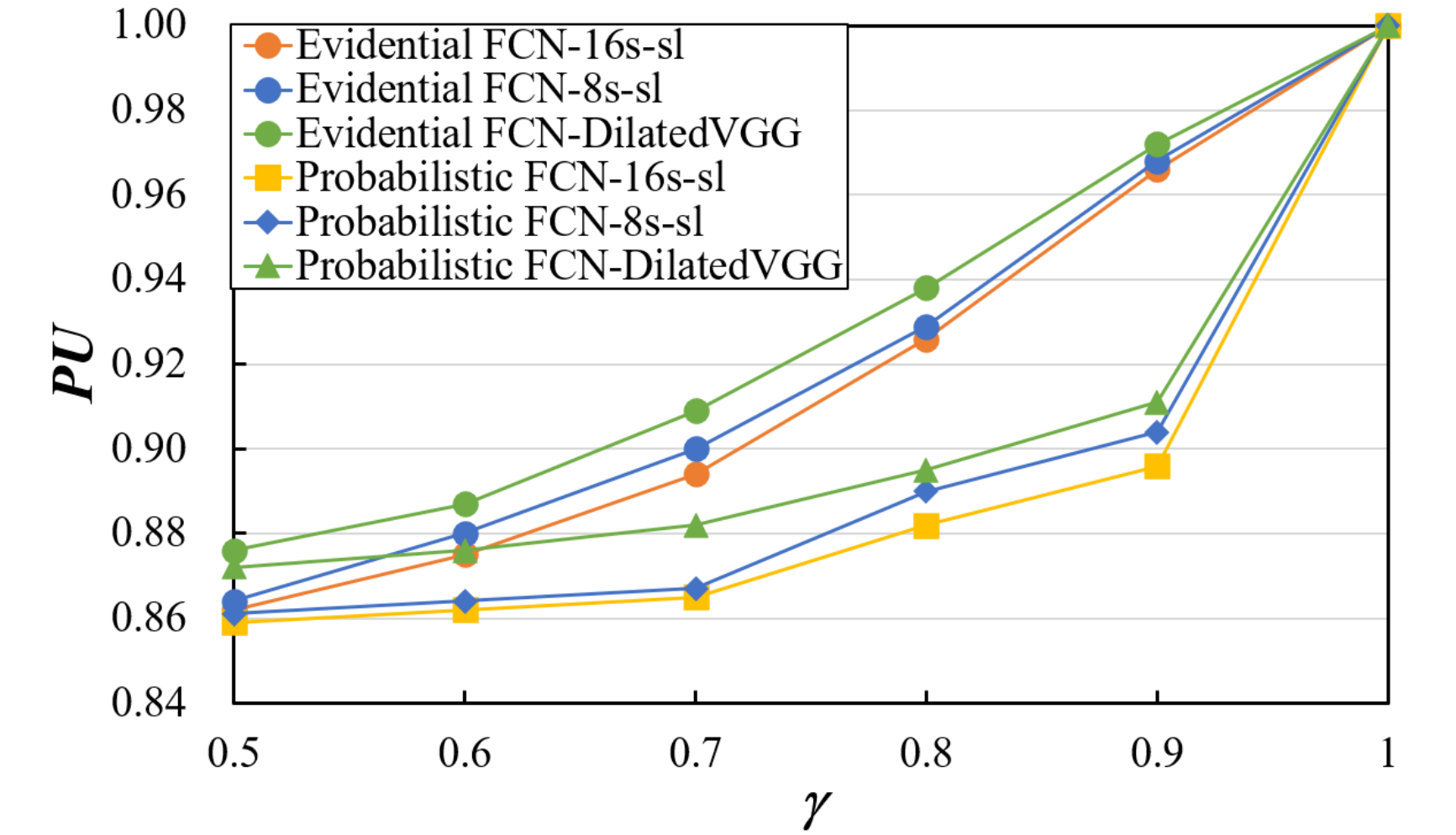}}\\
 \subfloat[\label{fig:uiou_nosoft_sift}]{\includegraphics[width=0.49\textwidth]{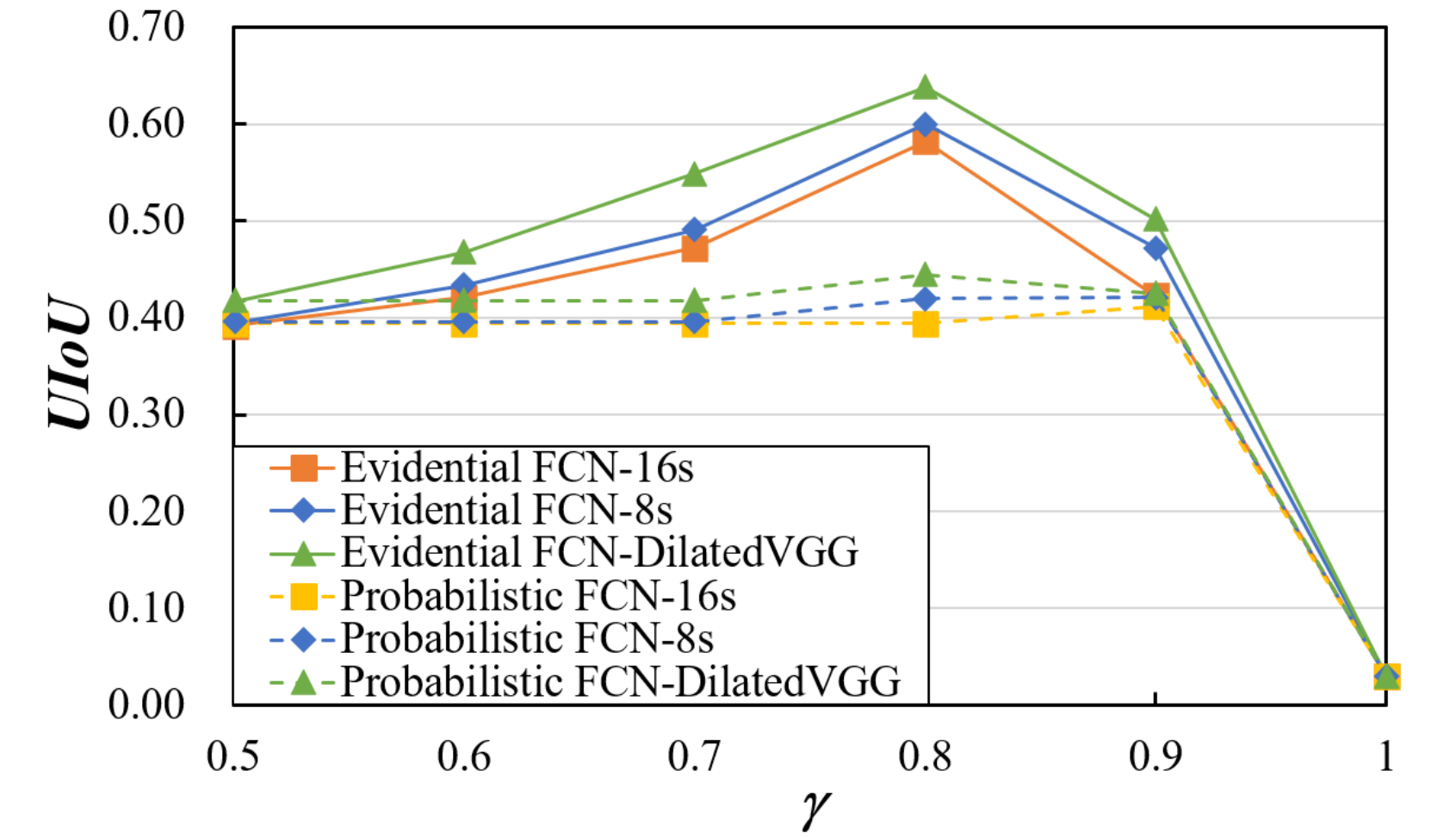}}\hspace{0.01\textwidth}
 \subfloat[\label{fig:uiou_soft_sift}]{\includegraphics[width=0.49\textwidth]{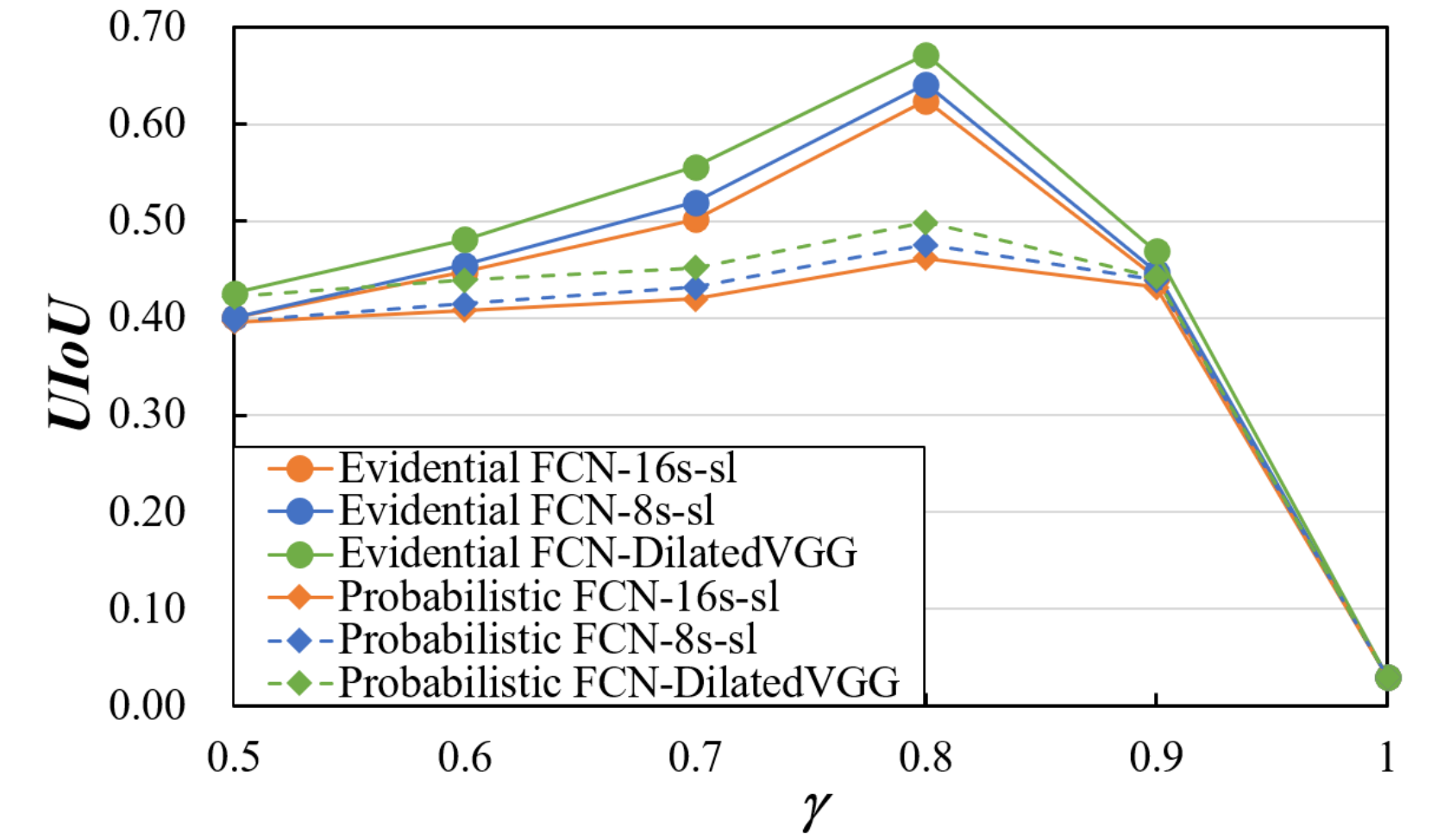}}\hspace{0.01\textwidth}
 \caption{Testing PU and UIoU vs. $\gamma$ on the SIFT Flow database. The first and second columns are the models trained with/without soft labels, respectively.}\label{fig:pu_uiou_imprecise_segmentation_sift}
\end{figure}

\begin{figure}
 \centering
 \includegraphics[width=0.85\textwidth]{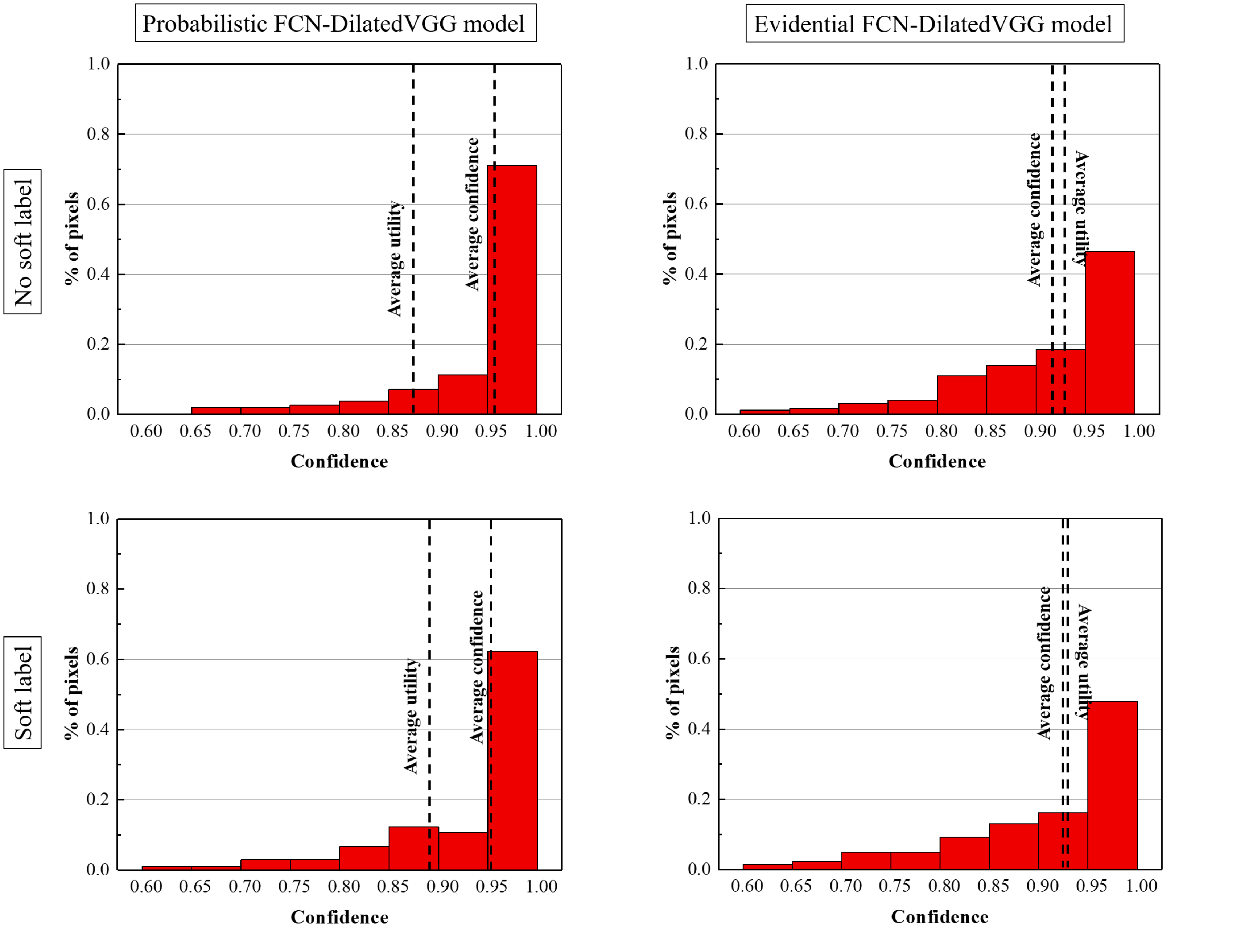}
 \caption{Pixel rate histograms for the P-FCN-DilatedVGG (left) and E-FCN-DilatedVGG (right) models on the SIFT Flow database without (top)/with (bottom) soft labels.}\label{fig:pixeldistribution_sift}
\end{figure}

Figure \ref{fig:pu_uiou_imprecise_segmentation_voc} displays the test results according to PU and UIoU for  imprecise segmentation of the Pascal VOC database. For a wide range of  imprecision tolerance degree $\gamma$, the E-FCN models reach higher PU and UIoU values than those obtained by the P-FCN models; this is due to the fact that the E-FCN models tend to assign  ambiguous pixels to multi-class sets, instead of making precise decisions. Such imprecise assignments avoid pixel-wise misclassification in case of high uncertainty, especially when feature vectors from an encoder-decoder architecture do not contain sufficient information to identify a precise class, and multiple classes have similar probabilities. Figure \ref{fig:pixeldistribution_voc} shows the pixel confidence distributions for the FCN models with $\gamma=0.8$. We can see that  the average confidences of the E-FCN models are smaller than those of the P-FCN models. This observation suggests that the E-FCN models make cautious decisions for ambiguous pixels by assigning them to multi-class sets, rather than classifying them arbitrarily into a single class. The E-FCN models are thus better calibrated than those based on P-FCN, which can be over-confident. Similar results are observed with the MIT-scene Parsing (Figures \ref{fig:pu_uiou_imprecise_segmentation_mit}-\ref{fig:pixeldistribution_mit}) and SIFT Flow  (Figures \ref{fig:pu_uiou_imprecise_segmentation_sift}-\ref{fig:pixeldistribution_sift}) databases.  We can thus conclude the  DS and utility layers improve the performance of the FCN models in  imprecise segmentation tasks by allowing us to assign some ambiguous pixels to multi-class sets.

\begin{figure}
 \centering
 \includegraphics[width=\textwidth]{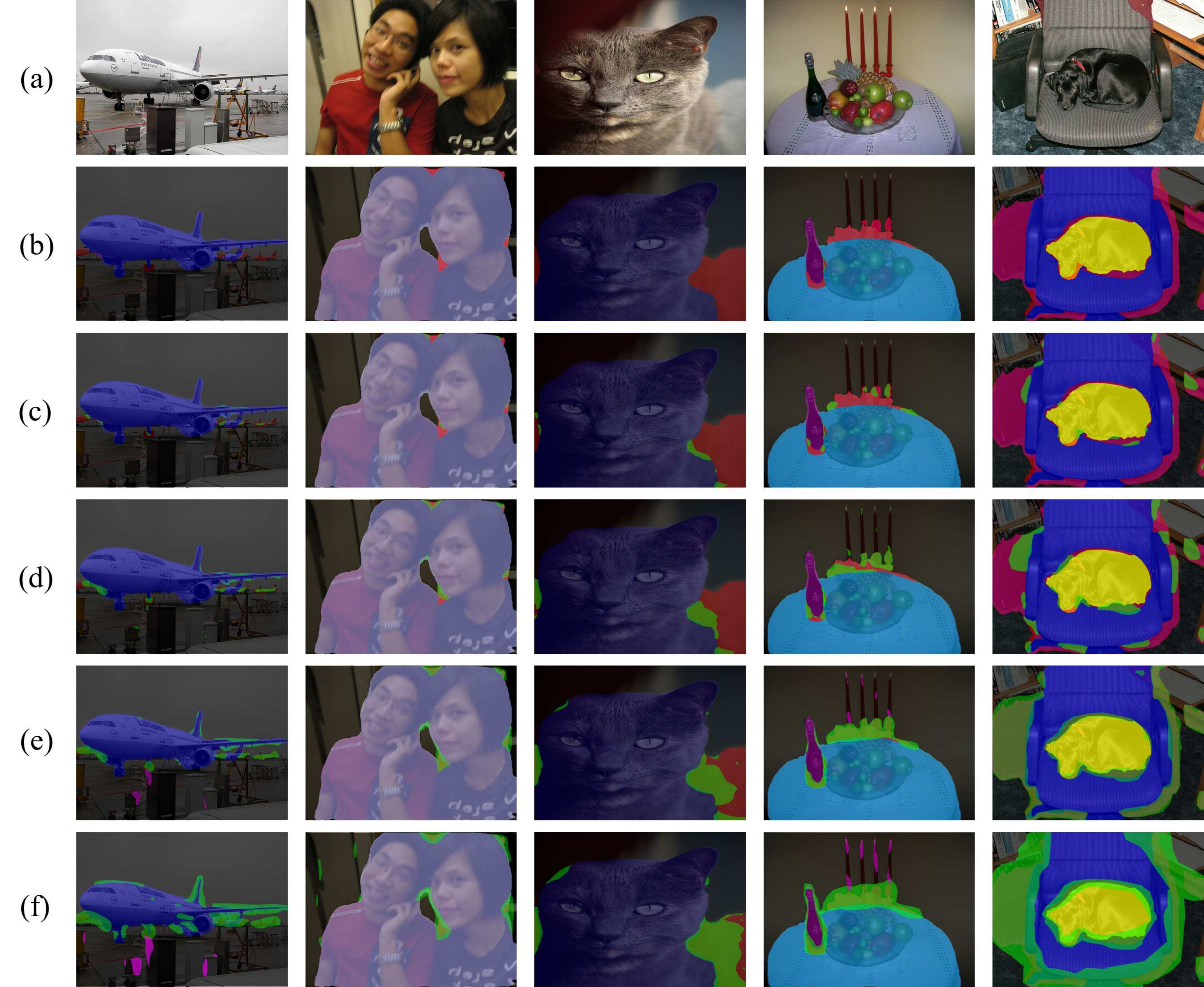}
 \caption{Segmentation examples from the Pascal VOC 2011 database: (a) Original image, (b) Precise segmentation, (c) Imprecise segmentation with $\gamma=0.6$, (d) Imprecise segmentation with $\gamma=0.7$,  (e) Imprecise segmentation with $\gamma=0.8$, and (f) Imprecise segmentation with $\gamma=0.9$. Red masks are pixels incorrectly classified in the precise segmentation; green masks are pixels assigned to multi-class sets except set $\Omega$; pink masks are pixels assigned to set $\Omega$;  other masks are pixels assigned to correct single-class sets. }\label{fig:segmentation_examples}
\end{figure}

In Figures \ref{fig:pu_uiou_imprecise_segmentation_voc}, \ref{fig:pu_uiou_imprecise_segmentation_mit} and  \ref{fig:pu_uiou_imprecise_segmentation_sift}, we can see that the value of UIoU first increases and then decreases when $\gamma$ increases from 0.5 to 1. To explain this behavior, Figure \ref{fig:segmentation_examples} illustrates some segmentation examples generated by the E-FCN-8s model trained on the Pascal VOC database with soft labels.  The first and second columns of  Figure \ref{fig:segmentation_examples} contain, respectively, the original images and their  precise segmentation predicted masks, while the third to sixth columns show the imprecise segmentation results for  values of $\gamma$ ranging  from 0.6 to 0.9. When $\gamma$ increases from 0.5 to 0.8, the majority of the green masks (the areas whose pixels are assigned to multi-class sets) tends to cover the red masks (the areas whose pixels are incorrectly classified in the precise segmentation). This observation can be explained by the fact that, in Eq. \eqref{con:uiou}, the increase in the utility of the intersection between predicted and labeled areas is larger than the increase in the union between the two areas. As a result, UIoU increases when $\gamma$ increases from 0.5 to 0.8. However, when $\gamma$ increases from 0.8 to 1.0, the majority of the green masks cover the areas predicted correctly in the precise segmentation, which causes the increase in the utility of intersection to be smaller than the increase in the union areas. This phenomenon leads to the decrease of UIoU when $\gamma$ is larger than 0.8.

\begin{figure}
 \centering
 \includegraphics[width=0.85\textwidth]{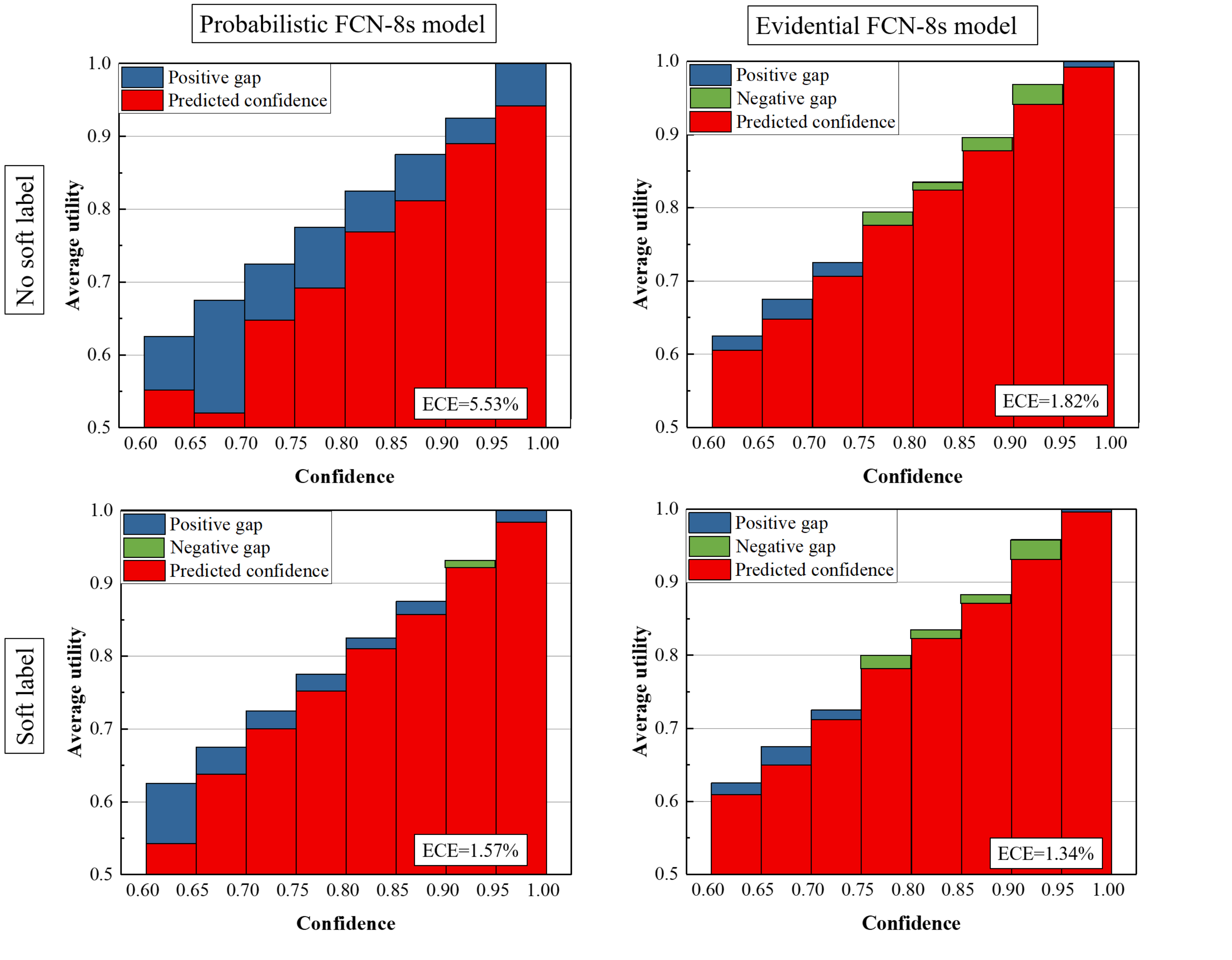}
 \caption{Average utility histograms for P-FCN-8s (left) and E-FCN-8s (right) with $\gamma=0.8$ on the Pascal VOC 2011 database without (top)/with (bottom) soft labels.}\label{fig:imprecision_ece_voc}
\end{figure}

\begin{figure}
 \centering
 \includegraphics[width=0.85\textwidth]{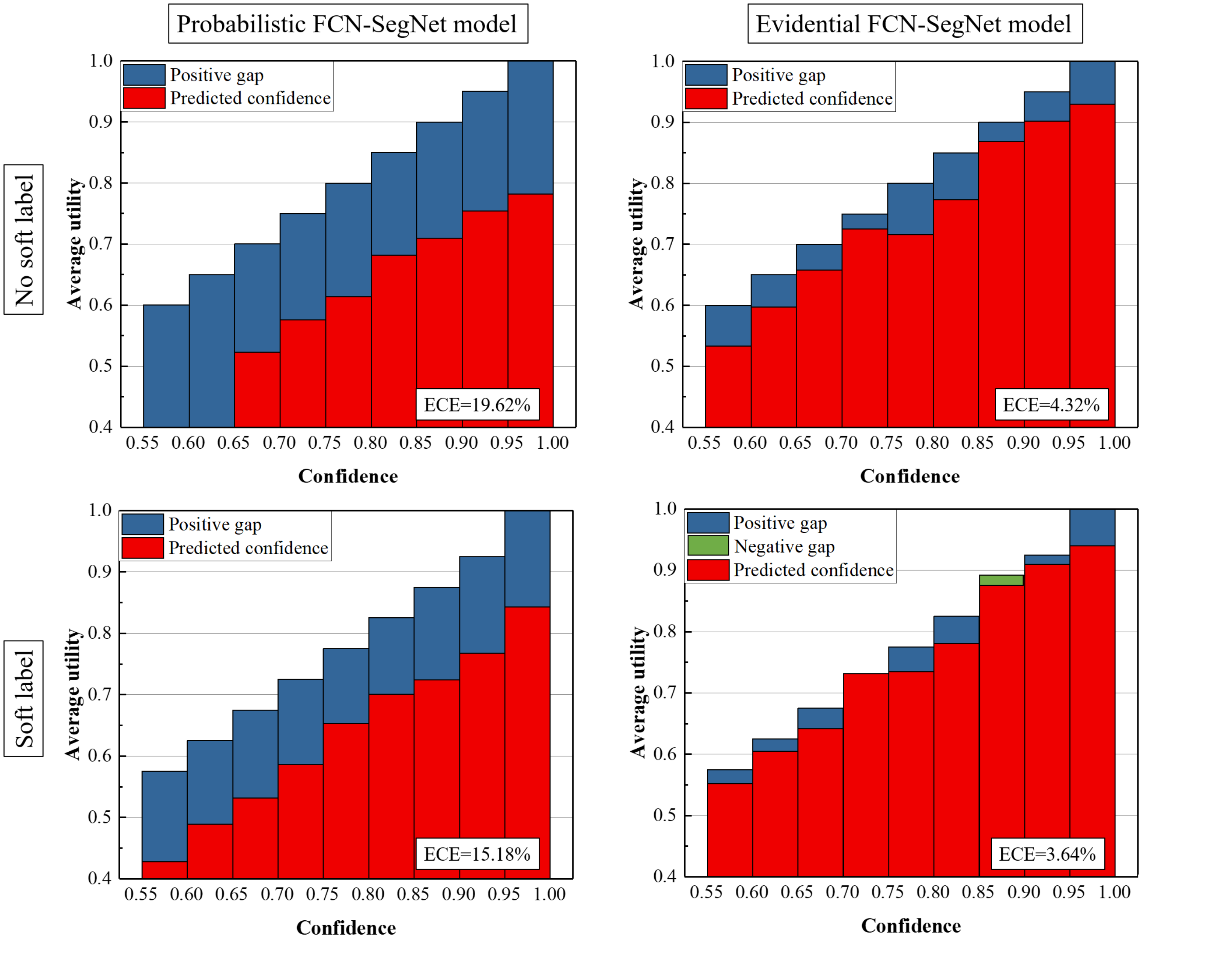}
 \caption{Average utility histograms for P-FCN-SegNet (left) and E-FCN-SegNet (right) with $\gamma=0.8$ on the MIT-scene Parsing database without (top)/with (bottom) soft labels.}\label{fig:imprecision_ece_mit}
\end{figure}

\begin{figure}
 \centering
 \includegraphics[width=0.85\textwidth]{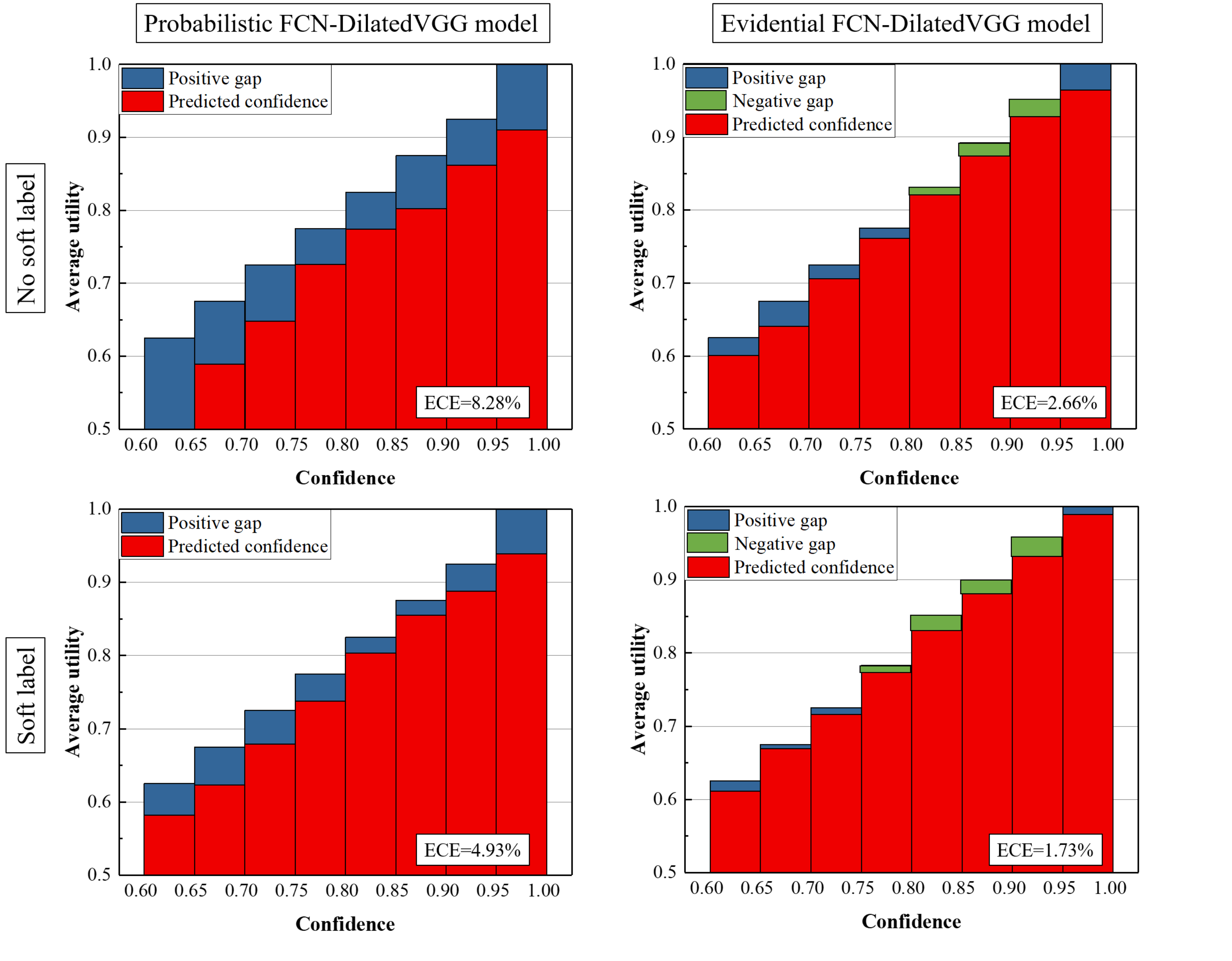}
 \caption{Average utility histograms for P-FCN-DilatedVGG (left) and E-FCN-DilatedVGG (right) with $\gamma=0.8$ on the SIFT Flow database without (top)/with (bottom) soft labels.}\label{fig:imprecision_ece_sift}
\end{figure}

The use of soft labels improves the performance of the FCN models for imprecision segmentation tasks. As shown in Figure \ref{fig:pu_uiou_imprecise_segmentation_voc}, the FCN models trained by the Pascal VOC database with soft labels have larger testing PU and UIoU than the ones without soft labels, which demonstrates the accuracy improvement using soft labels. Additionally, the use of soft labels can also improve the calibration of the FCN models. Figure \ref{fig:imprecision_ece_voc} shows that the ECEs and bin gaps in the E-FCN and P-FCN models are smaller when using the learning set with soft labels. These results demonstrates the feasibility of processing pixels with confusing information by using soft labels when training FCN models. The improvement of accuracy and calibration due to learning from soft labels can also be found with the MIT-scene Parsing and SIFT Flow databases, as shown, respectively, in Figures \ref{fig:imprecision_ece_mit} and \ref{fig:imprecision_ece_sift}. Therefore, we can conclude that the use of soft labels improves the accuracy and calibration of  FCN models.

\subsection{Novelty detection results}
\label{sec:novelty_detection}

\begin{figure}
 \centering
 \includegraphics[width=0.84\textwidth]{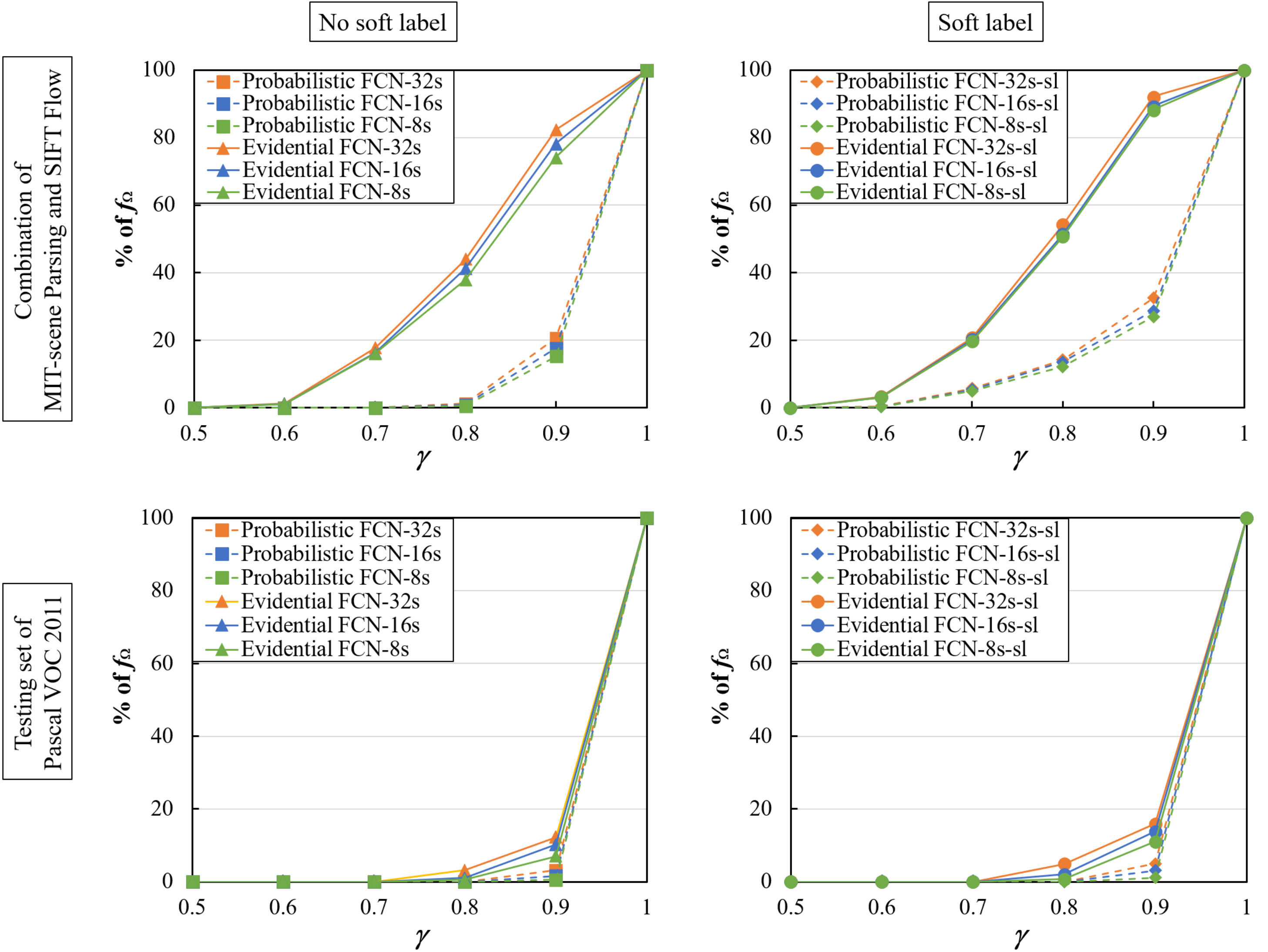}
 \caption{Proportion of pixels  assigned to $\Omega$ as a function of $\gamma$ for novelty detection on the combination of MIT-scene Parsing and SIFT Flow databases (top) and the testing set from the Pascal VOC 2011 database (bottom) when the learning set is from the Pascal VOC database without (left)/with (right) soft labels.}\label{fig:novelty_voc}
\end{figure}

\begin{figure}
 \centering
 \includegraphics[width=0.84\textwidth]{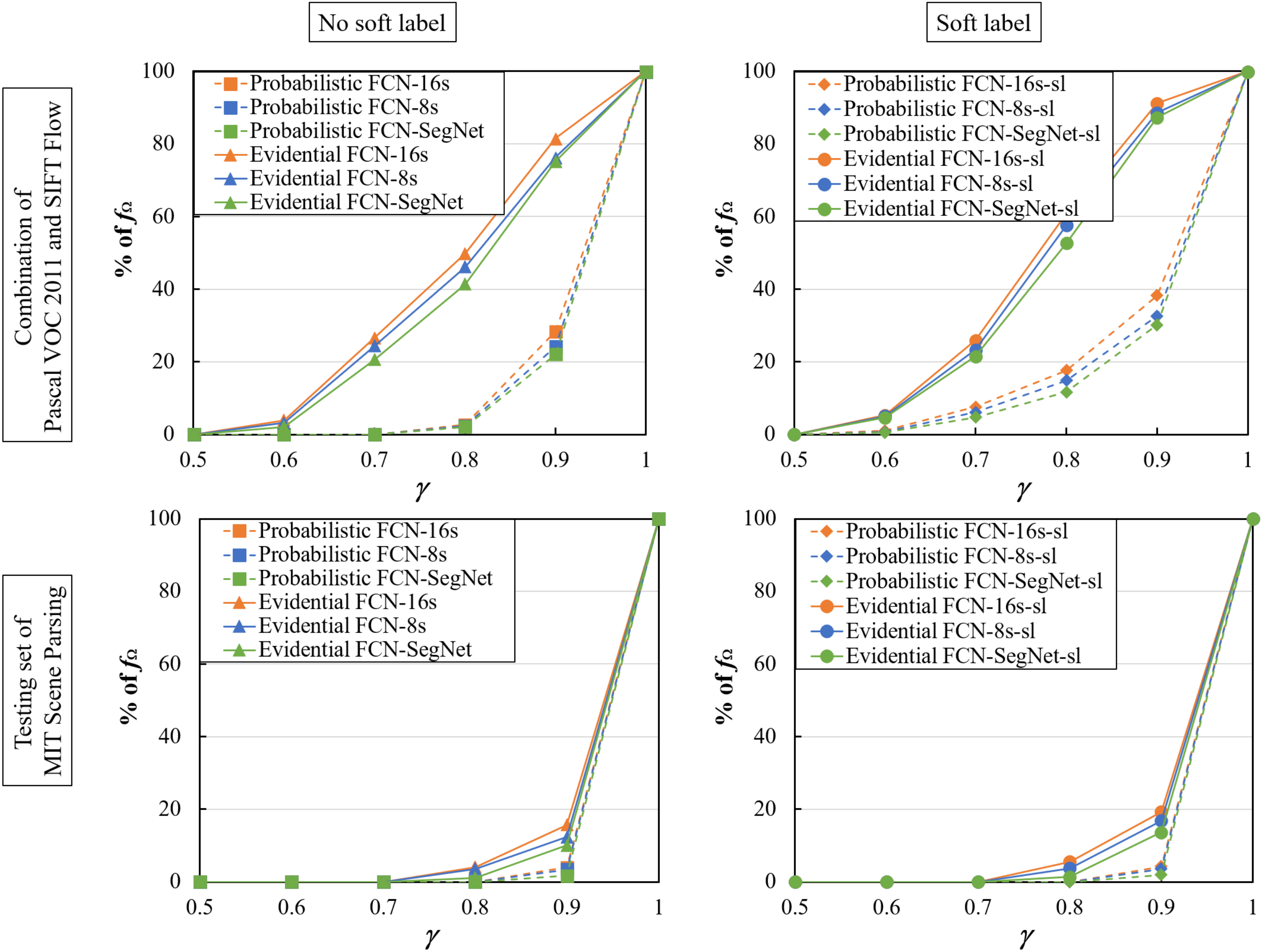}\\
 \caption{Proportion of pixels  assigned to $\Omega$ as a function of $\gamma$ for novelty detection on the combination of Pascal VOC 2011 and SIFT Flow (top) and the testing set of the MIT-scene Parsing database (bottom) when the learning set is from the Pascal VOC database without (left)/with (right) soft labels.}\label{fig:novelty_mit}
\end{figure}

\begin{figure}
 \centering
 \includegraphics[width=0.84\textwidth]{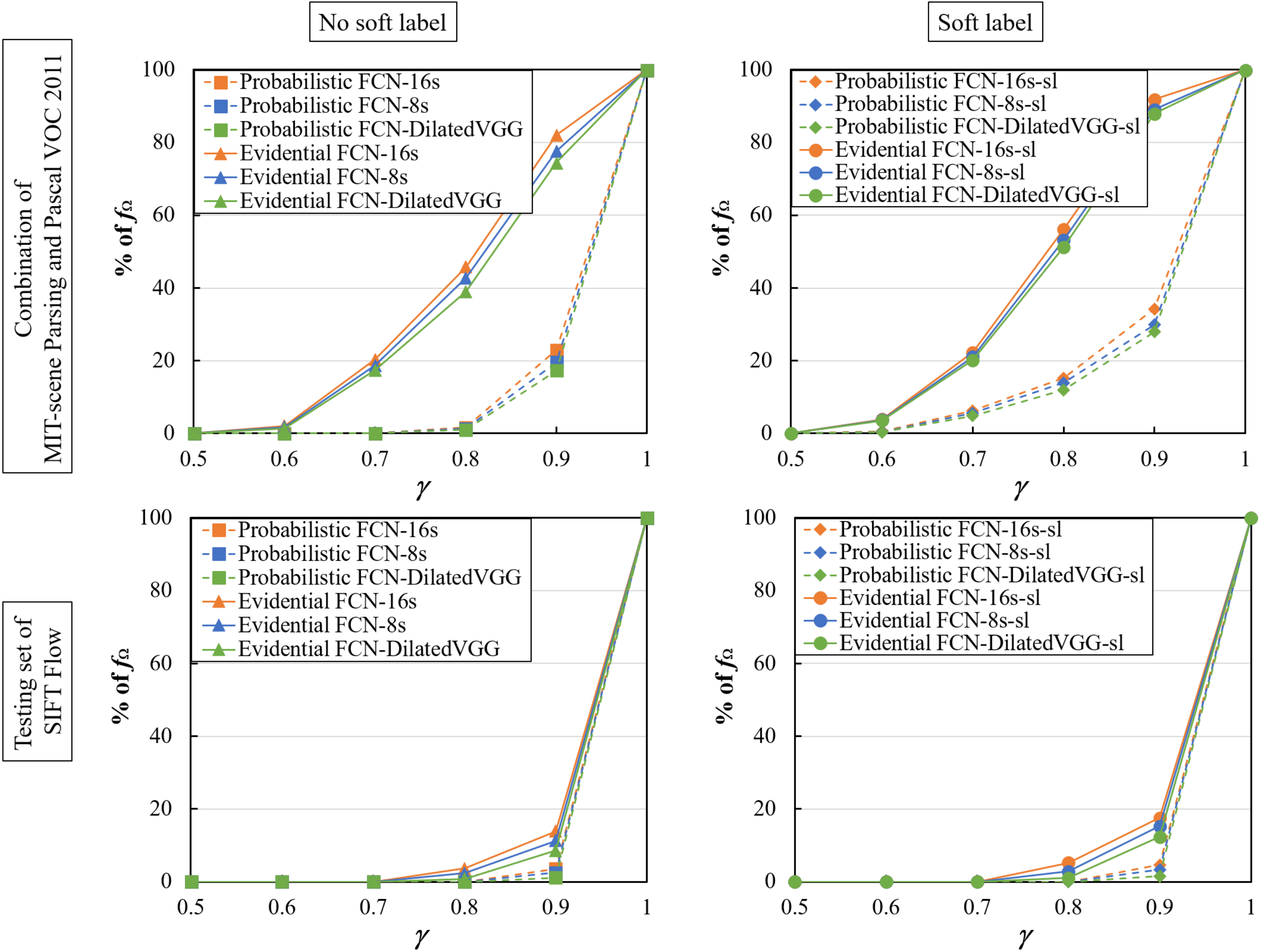}\\
 \caption{Proportion of pixels  assigned to $\Omega$ as a function of $\gamma$ for novelty detection on the combination of Pascal VOC 2011 and MIT-scene Parsing (top) and the testing set of the SIFT Flow database (bottom) when the learning set is from the Pascal VOC database without (left)/with (right) soft labels.}\label{fig:novelty_sift}
\end{figure}

For novelty detection, a pixel is considered as an outlier or an ambiguous sample if it is assigned to set $\Omega$. Figures \ref{fig:novelty_voc},  \ref{fig:novelty_mit} and \ref{fig:novelty_sift} show the results of novelty detection using the E-FCN and P-FCN models when the learning set is extracted, respectively, from the Pascal VOC, MIT-scene Parsing and SIFT Flow databases, and the test set is composed of images from the other two databases. In each testing set composed of two databases, only the pixels whose class is not represented in the corresponding learning set are reported in Figures \ref{fig:novelty_voc}-\ref{fig:novelty_sift}. The E-FCN models assign outliers and some known-class pixels to set $\Omega$ for values of $\gamma$  between 0.7 and 0.9, while the P-FCN models do not. This observation shows that the E-FCN models are more efficient than the probabilistic ones for rejecting outliers together with ambiguous samples. The proposed architecture thus has the potential to perform novelty detection once given a reasonable value of tolerance to imprecision. However, none of the FCN models performs well when $\gamma$ is less than 0.7 since these models favor precise decisions.

\begin{figure}
 \centering
 \includegraphics[width=\textwidth]{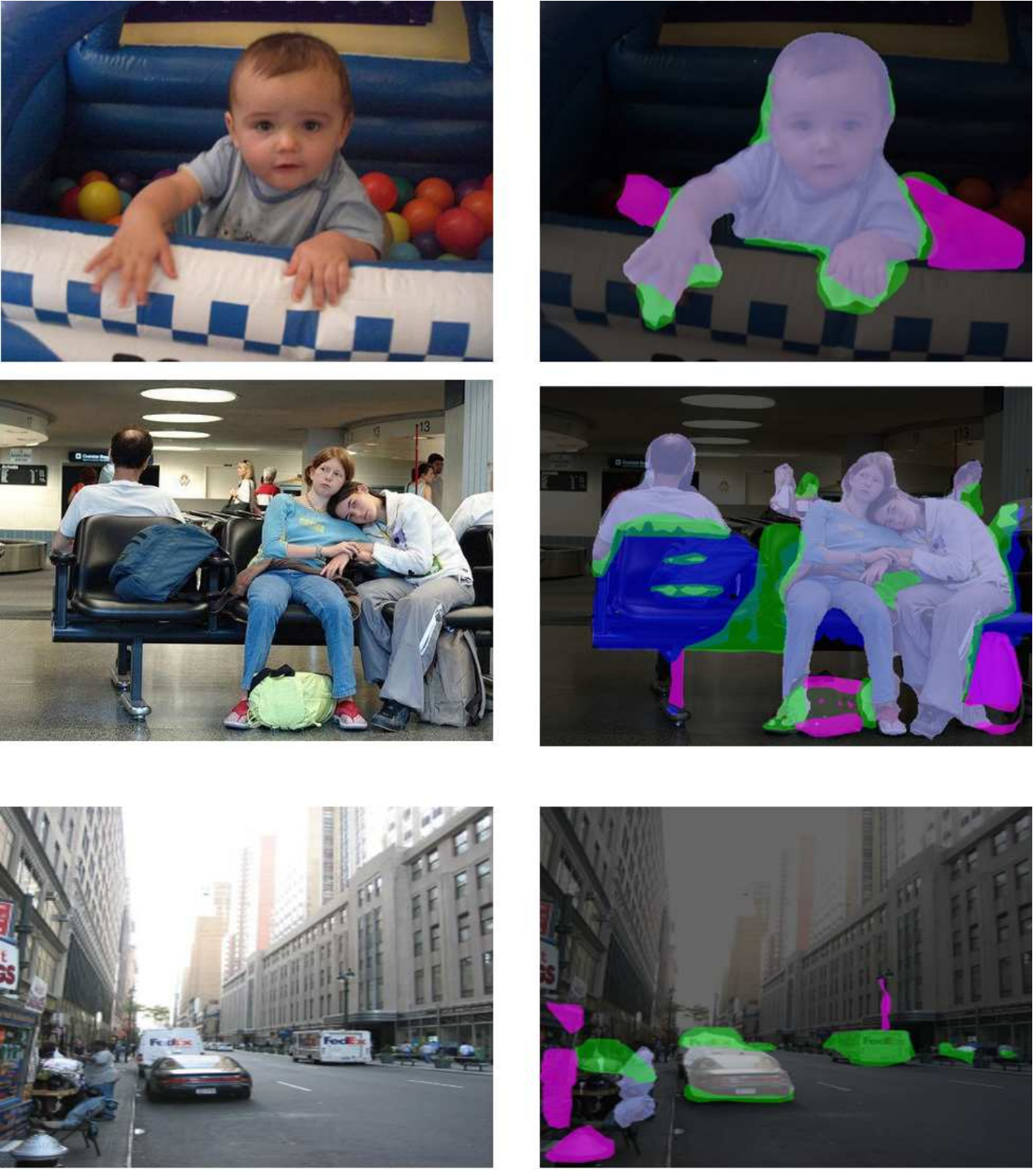}
 \caption{Examples of novelty detention from the MIT-scene Parsing database and their segmentation masks given by the E-FCN-8s model trained using the Pascal VOC database with soft labels when $\gamma$ equals 0.8. Red masks are pixels incorrectly assigned in the precise segmentation; green masks are pixels assigned to multi-class sets except set $\Omega$; pink masks are pixels assigned to set $\Omega$;  other masks are pixels assigned to correct single-class sets.}\label{fig:novelty_examples}
\end{figure}

The E-FCN models tend to reject unknown objects whose features are very different from those of the known objects in the learning set. For example, Figure \ref{fig:novelty_examples} shows images from the  MIT-scene Parsing database in which pixels representing  `bag', `street light' and `ball' objects are rejected by an E-FCN-8s model trained using the Pascal VOC database, which does not contain these objects. As shown  in  Table \ref{tab:novelty_example}, 75.2\% of the pixels representing a ball in the MIT-scene Parsing and and SIFT Flow databases are assigned to $\Omega$, while 16.1\% are assigned to a set of classes containing ``bottle''. For the ``bag'' and ``street light'' classes, these numbers are, respectively, 68.4\%/21.8\% and 77.3\%/16.3\%. Some unknown objects are not so easily rejected because of their similarity with known objects. For instance, 84.7\% of the pixels representing a seat and 81.7\% of pixels representing a bench are assigned to a set of classes containing ``chair'', and 88\% of ``wall'' pixels are assigned to a set of classes containing ``background''.

\begin{table}[]
	\centering
	\caption{Percentage of pixels from  some unknown classes in the MIT-scene Parsing and SIFT Flow databases classified by an E-FCN-8s model trained on the Pascal VOC database into some sets of classes. The model was trained with soft labels and $\gamma=0.8$.  For instance, 68.4\% of the pixels representing a bag were rejected (i.e., assigned to $\Omega$), and 84.7\% of pixels representing a seat were assigned to a set of classes containing the class ``chair''.}\label{tab:novelty_example}
	\begin{tabular}{clccccccc}
		\hline
		\multicolumn{2}{l}{\multirow{2}{*}{}}                                                      & \multicolumn{7}{c}{True class}                              \\ \cline{3-9} 
		\multicolumn{2}{l}{}                                                                       & bag  & street light & ball & seat & bench & bed  & wall \\ \hline
		\multirow{5}{*}{\begin{tabular}[c]{@{}c@{}}Assigned \\set\end{tabular}} & $\Omega$      & 68.4 & 77.3        & 75.2 & 7.8  & 4.7   & 15.9 & 4.9  \\
		& $\{\textrm{bottle}, \ldots\}$     & 21.8 & 16.3        & 16.1 & 48.5 & 39.7  & 30.3 & 0.2  \\
		& $\{\textrm{chair},\ldots \}$      & 11.3 & 9.2         & 8.5  & 84.7 & 81.7  & 58.6 & 0.3  \\
		&  $\{\textrm{background},\ldots\}$ & 15.2 & 13.7        & 11.5 & 58.7 & 48.6  & 46.9 & 88.0   \\
		& Others     & 4.2  & 2.4         & 1.5  & 2.7  & 3.5   & 5.2  & 3.7  \\ \hline
	\end{tabular}
\end{table}

We can also observe that the FCN models trained using a leaning set with soft labels reject more outliers than those trained without soft labels, as shown in Figures \ref{fig:novelty_voc}, \ref{fig:novelty_mit} and \ref{fig:novelty_sift}. This is because the use of soft labels makes the FCN models more cautious and better calibrated, as discussed in Section \ref{sec:imprecise_segmentation}. More precisely, for ambiguous pixels or outliers, the output mass functions of the FCN models trained with soft labels are more uniform than those computed by  FCN models trained without soft labels. As a result,  ambiguous pixels and outliers are more easily assigned to set $\Omega$. We can thus conclude that  soft labels have the potential to enhance  novelty detection performance.

\section{Conclusions}
\label{sec:conclusions}
In this paper, we have presented a new approach based on the combination of DS theory and  FCN for image semantic segmentation. In the proposed model,  called evidential fully convolutional network (E-FCN), an encoder-decoder architecture first extracts pixel-wise feature maps from an input image. A Dempster-Shafer layer then computes mass functions at each pixel location based on distances to prototypes.  Finally, a utility layer performs semantic segmentation based on pixel-wise mass functions. The proposed model can be trained using a learning set with soft labels in an end-to-end way.

The main finding of this study is that the proposed combination of FCNs and ENNs  makes it possible to improve accuracy and calibration of FCN models by assigning ambiguous pixels to multi-class sets, while maintaining the good performance of FCNs in precise segmentation tasks.  The E-FCN model is able to select a set of classes when the object representation does not allow us to select a single class unambiguously, which easily leads to incorrect decision-making in probabilistic FCNs. This result provides a new direction to improve the performance of FCN models for semantic segmentation.  The learning strategy using soft labels further improves the accuracy and calibration of the FCN models. Additionally, the proposed approach makes it possible to reject outliers together with ambiguous pixels when the tolerance to imprecision is between 0.7 and 0.9.

Future work will focus on two main aspects. First, we will investigate multi-FCN-model information fusion for semantic segmentation based on definition of soft labels, using an approach similar to that introduced  in \cite{xu2016multimodal}. Other advanced evidential classifiers, such as the contextual-discounting evidential $K$-nearest neighbor \cite{denoeux19f} will also be considered to improve the performance of the proposed neural network architecture.


%
%


\bibliographystyle{spmpsci}      
\bibliography{mybibfile}
\end{document}